\documentclass[10pt,twocolumn,letterpaper]{article}

\usepackage[pagenumbers]{wacv} % To force page numbers, e.g. for an arXiv version
\definecolor{wacvblue}{rgb}{0.21,0.49,0.74}
\usepackage[pagebackref,breaklinks,colorlinks,allcolors=wacvblue]{hyperref}

\def\wacvPaperID{0436} % *** Enter the WACV Paper ID here
\def\confName{WACV}
\def\confYear{2026}

\usepackage{multirow}
\usepackage{multicol}
\usepackage{colortbl}
\usepackage{xcolor}
\usepackage{pifont}
\usepackage{makecell}
\usepackage{tabularx}
\usepackage{tikz}

\newcommand{\circled}[1]{\tikz[baseline=(char.base)]{
            \node[shape=circle,draw,inner sep=1pt] (char) {\scriptsize #1};}}

\title{AnyAnomaly: Zero-Shot Customizable Video Anomaly Detection with LVLM}

\author{
    Sunghyun Ahn\thanks{Equal contribution},
    \ Youngwan Jo\footnotemark[1],
    \ Kijung Lee,
    \ Sein Kwon,
    \ Inpyo Hong,
    \ Sanghyun Park\thanks{Corresponding author: Sanghyun Park (sanghyun@yonsei.ac.kr)} \\
    Yonsei University, Seoul, Republic of Korea \\
    {\tt\small \{skd, jyy1551, rlwjd4177, seinkwon97, hip9863, sanghyun\}@yonsei.ac.kr}
}

\begin{document}
\maketitle
\begin{abstract}
Video anomaly detection (VAD) is crucial for video analysis and surveillance in computer vision. However, existing VAD models rely on learned normal patterns, which makes them difficult to apply to diverse environments. Consequently, users should retrain models or develop separate AI models for new environments, which requires expertise in machine learning, high-performance hardware, and extensive data collection, limiting the practical usability of VAD. To address these challenges, this study proposes customizable video anomaly detection (C-VAD) technique and the AnyAnomaly model. C-VAD considers user-defined text as an abnormal event and detects frames containing a specified event in a video. We effectively implemented AnyAnomaly using a context-aware visual question answering without fine-tuning the large vision language model. To validate the effectiveness of the proposed model, we constructed C-VAD datasets and demonstrated the superiority of AnyAnomaly. Furthermore, our approach showed competitive results on VAD benchmarks, achieving state-of-the-art performance on UBnormal and UCF-Crime and surpassing other methods in generalization across all datasets. Our code is available online at \href{https://github.com/SkiddieAhn/Paper-AnyAnomaly}{\textcolor{wacvblue}{github.com/SkiddieAhn/Paper-AnyAnomaly}}.
\end{abstract}

\section{Introduction}
\label{sec:intro}
Video anomaly detection (VAD) aims to detect abnormal events in video streams. Abnormal events include the actions of objects that are inappropriate for the environment (e.g., climbing over a fence) or objects with unusual appearances (e.g., a bicycle on a walkway). However, abnormal events are rare and diverse, making it difficult to construct large-scale datasets for VAD. Therefore, VAD is recognized as a highly challenging problem.

To overcome these limitations, previous studies primarily used one-class classification (OCC) methods that learn only from normal data. In the OCC approach, the model learns normal patterns and classifies the cases that deviate from them as abnormal. Representative OCC methods include classification-\cite{ruff2018deep, yi2020patch, lee2020multi}, distance-\cite{roth2022towards, ahn2024videopatchcore, reissattribute}, and prediction-based methods \cite{liu2018future, yang2023video, hong2024making, lee2025mdvad}, all of which have demonstrated excellent performance in VAD tasks.

\begin{figure}[t]
  \centering
   \includegraphics[width=\linewidth]{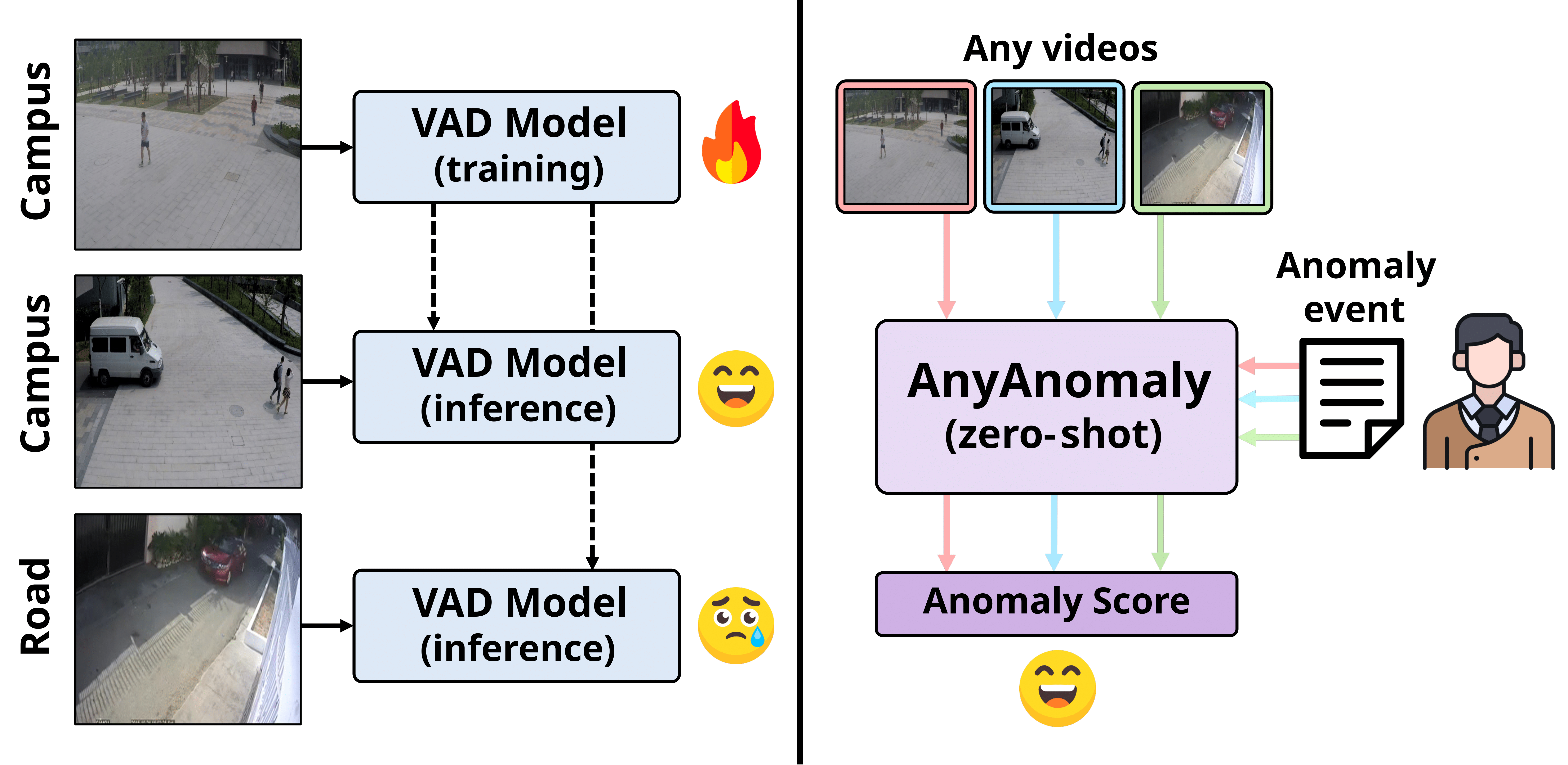}
   \caption{Comparison of traditional video Anomaly Detection (VAD) and customizable video anomaly detection (C-VAD). Traditional VAD models struggle with generalization, making them hard to apply in diverse environments, while C-VAD can handle various video environments.}
   \label{fig1}
\end{figure}

However, because normal and abnormal classes can be defined differently depending on the environment, OCC methods cannot always guarantee a generalized performance. For example, as shown on the left side of \Cref{fig1}, a model trained in a campus environment learns the characteristics of a `person’ as a normal pattern and classifies a `car’ as abnormal. However, when this model is applied to a road environment, a `car’ is still detected as abnormal, which can increase the number of false positives. Therefore, OCC methods require the retraining of normal patterns for each new environment, which entails additional costs, such as data collection, expert intervention, and high-performance equipment. Because of these limitations, the application of VAD models to real-world scenarios is challenging.  

To address this issue, we propose a novel technique called customizable video anomaly detection (C-VAD). C-VAD considers user-defined text as abnormal events and detects the frames containing these events in the video. For instance, in campus videos, ‘car’ can be set as an abnormal event, while in road videos, ‘person’ can be set as an abnormal event. In contrast to existing VAD models, which judge abnormalities based on learned normal patterns, C-VAD dynamically detects abnormal patterns based on the text provided. This implies that as the generalizability of visual text analysis improves, anomaly detection becomes more effective in various environments. Consequently, we introduce a zero-shot capable C-VAD approach, as shown on the right side of \Cref{fig1}, and propose the AnyAnomaly model, which allows for VAD in various environments without the need for additional training.  

An effective method to implement zero-shot C-VAD is to leverage large vision language models (LVLMs). Recently, LVLMs have demonstrated outstanding generalization performance in visual text analysis. By leveraging this capability, C-VAD can be performed effectively across various environments. The most intuitive method involves performing visual question answering (VQA) \cite{antol2015vqa} on each frame to estimate the anomaly score. For instance, one could provide the model with the prompt: \textit{“Return a value between 0 (no) and 1 (yes) indicating how well the input image represents the text provided by the user”}. This was used as the baseline model. However, through experiments, we observed the following limitations of the baseline model: \circled{1} Due to the large computational cost of LVLMs, the latency is high. \circled{2} Difficulty in analyzing specific objects due to the characteristics of surveillance videos, such as foreground-background imbalance and object congestion. \circled{3} Difficulty in detecting action-related anomalies because of the inability to utilize temporal information.  

\begin{figure}[t]
  \centering
   \includegraphics[width=\linewidth]{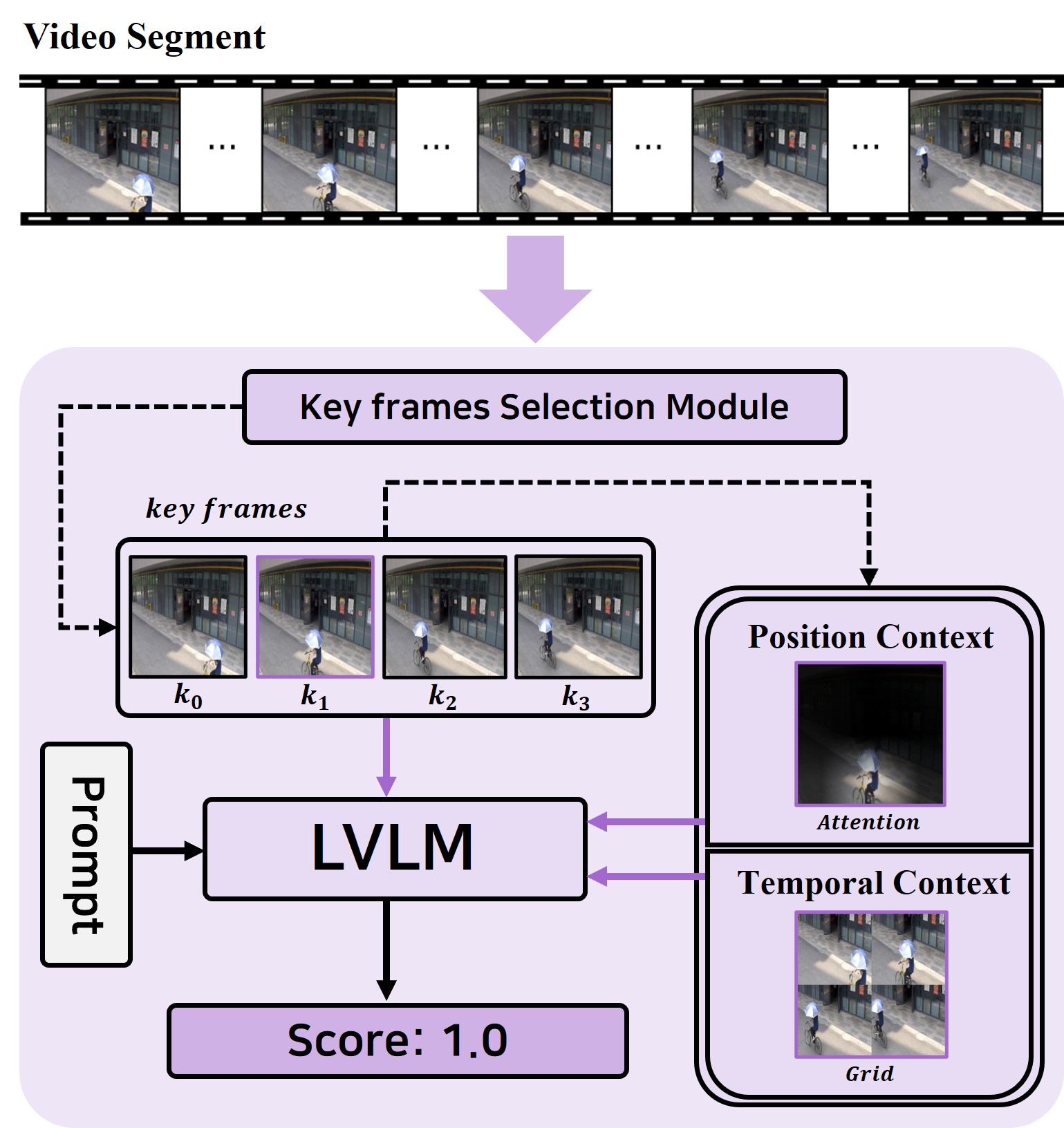}

   \caption{The architecture of AnyAnomaly}
   \label{fig2} 
\end{figure}

To overcome these limitations, we designed an AnyAnomaly model with the structure shown in \Cref{fig2}. First, to reduce the latency, we adopted a segment-level approach that groups consecutive frames into a single segment for processing. For this purpose, we introduced a key frames selection module (KSM) that selects key frames representing the segment and performed VQA per segment. Second, instead of performing simple image-text matching, we introduced a context-aware VQA approach to enable a deeper understanding of the scene. To this end, we additionally utilized two types of information: position context, $PC$ and temporal context, $TC$. $PC$ is a context that emphasizes important locations within a frame, enhancing the object analysis capability of the LVLM. $TC$ is a context that structures scene changes over time into a grid format, improving the action analysis capability of the LVLM. Notably, the proposed KSM and context generation modules operate in a training-free manner, allowing for easy application of C-VAD without additional training.  

To evaluate the performance of C-VAD, we classified existing VAD benchmark datasets based on anomaly types to create the C-VAD datasets. Through this process, we demonstrated the superiority of AnyAnomaly. Furthermore, despite being a zero-shot approach, AnyAnomaly achieved competitive performance on VAD datasets compared to traditional OCC-based VAD models. Remarkably, It achieved state-of-the-art (SOTA) results on UBnormal \cite{acsintoae2022ubnormal} and UCF-Crime \cite{sultani2018real}, and showed superior generalization across all datasets. The proposed approach is expected to be an effective solution for deploying VAD technology in real-world applications. The contributions of this study are as follows:
\begin{itemize}
    \item We propose the C-VAD technique for anomaly detection in diverse environments. To the best of our knowledge, it is the first to perform VAD based on user-defined anomalies.
    \item We develop the AnyAnomaly model, which applies context-aware VQA to perform C-VAD effectively.  
    \item To evaluate the performance of C-VAD, we construct new C-VAD datasets and experimentally verify the superiority of AnyAnomaly.
    \item AnyAnomaly achieves SOTA performance on UBnormal and UCF-Crime and outperforms other methods in generalization across all datasets.
\end{itemize}

\section{Related Work}
\noindent\textbf{Video Anomaly Detection.}
Most VAD models adopt the OCC approach to detect anomalies by learning normal patterns. Among them, prediction-based methods train models to predict future or past frames based on normal frames, assuming that abnormal frames exhibit larger prediction errors. Liu \cite{liu2018future} proposed a method that utilizes FlowNet \cite{ilg2017flownet} and GANs \cite{mao2017least} to predict the $t+1$-th frame given $t$ input frames. Yang \cite{yang2023video} introduced an approach that selects key frames from $t$ input frames to predict the entire sequence. However, because the definitions of normal and abnormal patterns may vary depending on the environment, the OCC approach, which relies on learned normal patterns, has a limited generalization performance. To mitigate this limitation, recent studies have explored cross-domain VAD (xVAD). Both rGAN \cite{lu2020few} and MPN \cite{lv2021learning} adopted meta-learning approaches, enabling adaptation to a target domain from only a few training samples. zxVAD \cite{aich2023cross} further improved adaptability to new environments by synthesizing abnormal patterns through the cut-mix technique applied to images from auxiliary datasets without explicit adaptation to the target domain. However, this approach depends on fixed data transformations, making it difficult to fully capture the diverse abnormal patterns that may occur in real-world scenarios. Therefore, we propose a novel VAD method that uses textual information to dynamically detect abnormal patterns that vary depending on the environment.\\

\noindent\textbf{Large Vision Language Models.}
Large language models have primarily been used in natural language processing; however, they have recently been applied to multimodal tasks such as image captioning and VQA. For example, MiniGPT-4 \cite{zhu2023minigpt} processes multimodal inputs by connecting a pre-trained vision encoder to the Vicuna \cite{chiang2023vicuna} model through a linear layer. Recent LVLMs have employed novel visual encoding techniques to better understand images. Chat-UniVi \cite{jin2024chat} generates dynamic tokens for images, thereby reducing unnecessary information and effectively extracting key visual features. This model enables flexible analysis by applying dynamic tokens across various resolutions. MiniCPM-V \cite{yao2024minicpm} applies the best partition technique according to the image resolution and generates tokens optimized for each segment, thereby improving the memory efficiency. However, despite the advancements in LVLMs, they are trained for general purposes, making their direct application in VAD challenging. Therefore, we propose a training-free approach to minimize the domain gap between the LVLMs and VAD tasks.
\section{Method}

\subsection{Overview}
\Cref{fig2} illustrates the structure of the AnyAnomaly model, which performs context-aware VQA. The input is a video segment $S = \{s_0, \dots, s_{N-1}\}$ comprising $N$ frames, where $N$ is a multiple of $4$. The KSM selects key frames $K = \{k_0, \dots, k_3\}$ from $S$. Among the selected key frames, the representative frame $\hat{k}$ is used to generate PC, whereas $K$ is used to create $TC$. Subsequently, $\hat{k}$, $PC$, and $TC$ are utilized as image inputs for the LVLM, whereas the user-provided text $X$ is combined with a prompt and used as the text input. Finally, the LVLM’s response results are integrated to compute an anomaly score.

The user-defined text $X$ refers to a natural language description of the anomaly the user wishes to detect. It can be a single word (e.g., “bicycle”), diverse events (e.g., “jumping-falling”), or a complex behavior (e.g., “driving outside the lane”). In the case of diverse events, each event keyword is processed individually as a single word.

\subsection{Key frames Selection Module}
\begin{figure*}[t]
    \centering
    \begin{subfigure}[h]{0.33\textwidth}
        \centering
        \includegraphics[width=\textwidth]{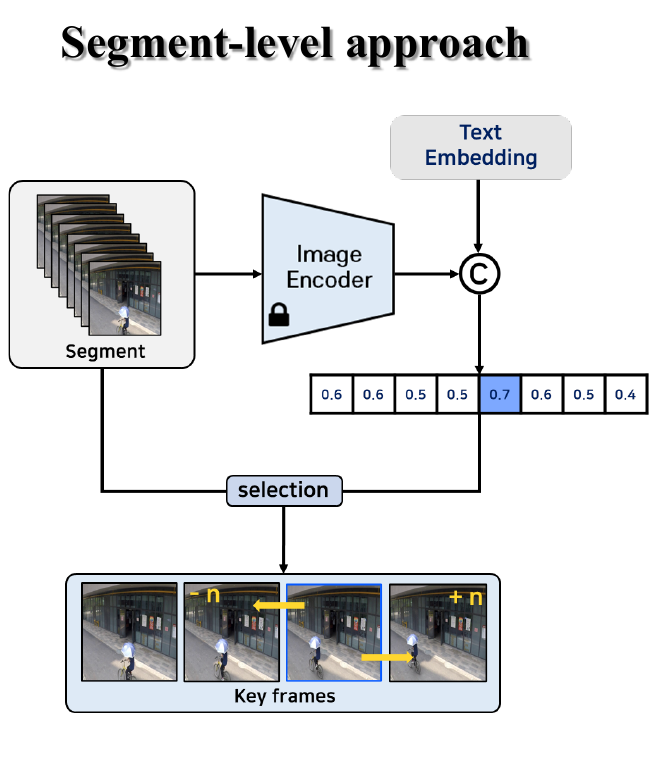}
        \caption{Key frames Selection Module (KSM)}
        \label{fig:figure3a}
    \end{subfigure}
    \hfill
    \begin{subfigure}[h]{0.33\textwidth}
        \centering
        \includegraphics[width=\textwidth]{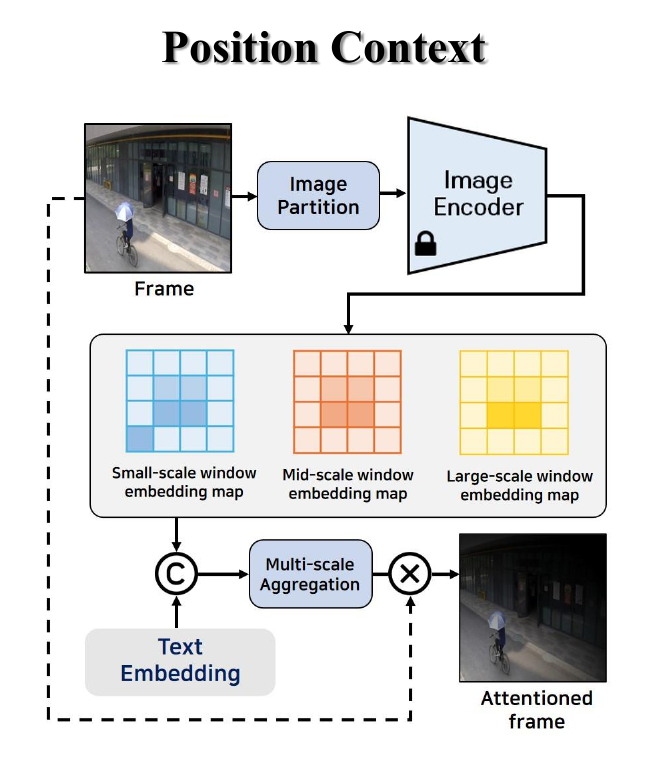}
        \caption{WinCLIP-based Attention (WA)}
        \label{fig:figure3b}
    \end{subfigure}
    \hfill
    \begin{subfigure}[h]{0.33\textwidth}
        \centering
        \includegraphics[width=\textwidth]{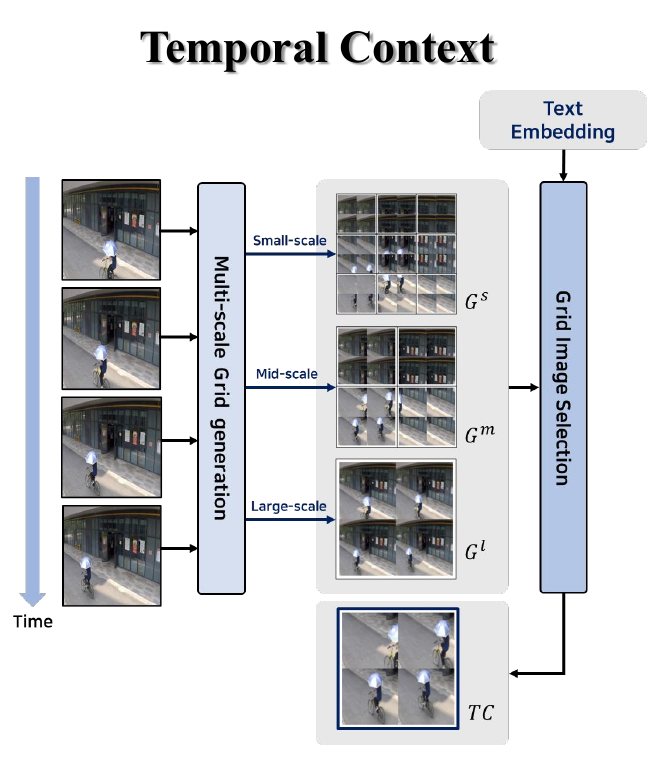}
        \caption{Grid Image Generation (GIG)}
        \label{fig:figure3c}
    \end{subfigure}
    \caption{Architecture of the proposed modules. KSM is essential for the segment-level approach, and WA and GIG are crucial for context generation.}
    \label{fig3}
\end{figure*}

\Cref{fig:figure3a} shows KSM, a key component of the segment-level approach. For this purpose, we selected four frames representing the segment as $K$ and utilized the CLIP \cite{radford2021learning} model, which was trained to match the images and text.

Specifically, $S$ and $X$ are inputs to the image encoder $E_I$ and text encoder $E_T$, respectively, and the similarity is calculated using the dot product of $N$ image and text embeddings. The frame with the highest similarity is selected as the representative frame $\hat{k}$.
\begin{equation}
\hat{k} = \underset{s_i \in S}{\arg\max} \left( E_I(s_i) \cdot E_T(X) \right)
\end{equation}
The index of the representative frame $\hat{k}$, denoted as $\hat{i}$, is used to select the other key frames. We divide the segment into four groups of equal size and select the $\hat{i} \bmod \frac{N}{4}$-th frame from each group. For example, when $N=8$ and $\hat{i}=4$, the $0$-th frame from each group is selected and the final set is $K = \{s_0, s_2, s_4, s_6\}$. This process is defined as follows:
\begin{equation}
k_i = s_{x}, \quad x = (i \times \frac{N}{4}) + (\hat{i} \bmod \frac{N}{4})
\end{equation}
Using the KSM, $K$ is generated by considering both text alignment and temporal uniformity, thereby enabling effective context generation. A comparative analysis of the key frames selection method is presented in \Cref{sec:ablation_study}.

\subsection{Context Generation}
$PC$ and $TC$ are key elements of context-aware VQA, serving as additional information that complements the input image. $PC$ enhances the object analysis capability of the LVLM and is generated through WinCLIP-based attention (WA). $TC$ strengthens the action analysis ability of the LVLM and is created through grid image generation (GIG).\\

\noindent\textbf{WinCLIP-based Attention.}
\Cref{fig:figure3b} illustrates the WA method. We emphasize the regions related to $X$ at $\hat{k}$ based on WinCLIP, as proposed by Jeong \cite{jeong2023winclip}. First, $\hat{k}$ is divided into multiple windows, and individual embeddings are generated from each window using $E_I$. For example, when the image size is $240 \times 240$, it is divided into 25 windows of size $48 \times 48$, and the embeddings of each window are collected to create a small-scale window embedding map $W^s\in \mathbb{R}^{25 \times D}$. By adjusting the window size, a middle-scale window embedding map $W^m$ and large-scale window embedding map $W^l$ are also generated, and the similarity between these embedding maps and the text embedding $z\in \mathbb{R}^{D}$ is calculated. The final similarity map $M$ is generated by averaging the similarities calculated on three scales:
\begin{equation}
M = \frac{1}{3}(z(W^s)^T + z(W^m)^T + z(W^l)^T)
\end{equation}
We combined the template proposed by Jeong \cite{jeong2023winclip} with $X$ and passed it through $E_T$ to generate $z$. Finally, we multiplied $M$ and $\hat{k}$ to create $PC$:
\begin{equation}
PC = f_{\text{norm}}(M) \odot \hat{k}
\end{equation}
Here, $f_{\text{norm}}$ represents min-max normalization, and $\odot$ denotes element-wise multiplication. $M$ was used after interpolation and reshaping to match the resolution of $\hat{k}$. Because $PC$ is created by integrating similarities from multiple scales, it is robust to object size and location, and operates effectively even in situations with multiple objects.\\

\noindent\textbf{Grid Image Generation.}
\Cref{fig:figure3c} illustrates the GIG method, which comprises two stages. In the multiscale grid generation stage, $K$ is used to create grid images at different scales. Similar to the process described in WA, each frame of $K$ is divided into multiple windows, and the windows at the same position are connected in a 2$\times$2 grid format to create a single grid image. This process is defined as follows:
\begin{equation}
g^i = \begin{bmatrix} 
u_0^i & u_1^i \\ 
u_2^i & u_3^i 
\end{bmatrix}
\end{equation}
Here, $u_j^i$ refers to the $i$-th window created from $k_j$, and $g^i$ refers to the $i$-th grid image. We defined the sets of grid images generated using small-, middle-, and large-scale windows as $G^s$, $G^m$, and $G^l$, respectively.
 
In the grid image selection stage, the previously created sets are aggregated to generate $G^{all}$. Then, using the same method as in KSM to select $\hat{k}$, the grid image with the highest similarity to the text is chosen to generate $TC$:
\begin{equation} 
G^{all} = G^s \cup G^m \cup G^l 
\end{equation}
\begin{equation} 
TC = \underset{g^i \in G^{all}}{\arg\max} \left( E_I(g^i) \cdot E_T(X) \right)
\end{equation}
The $TC$ generated through this process represents the object movement over time within the same background, making it advantageous for action analysis and robust to various object sizes. An analysis of the window sizes used in the WA and GIG is presented in \Cref{sec:ablation_study}.

\subsection{Anomaly Detection}
Instead of tuning the LVLM, we propose a new prompt and context for performing context-aware VQA. The VQA results were used as anomaly scores to enable training-free zero-shot anomaly detection.\\

\begin{figure}[t]
  \centering
   \includegraphics[width=\linewidth]{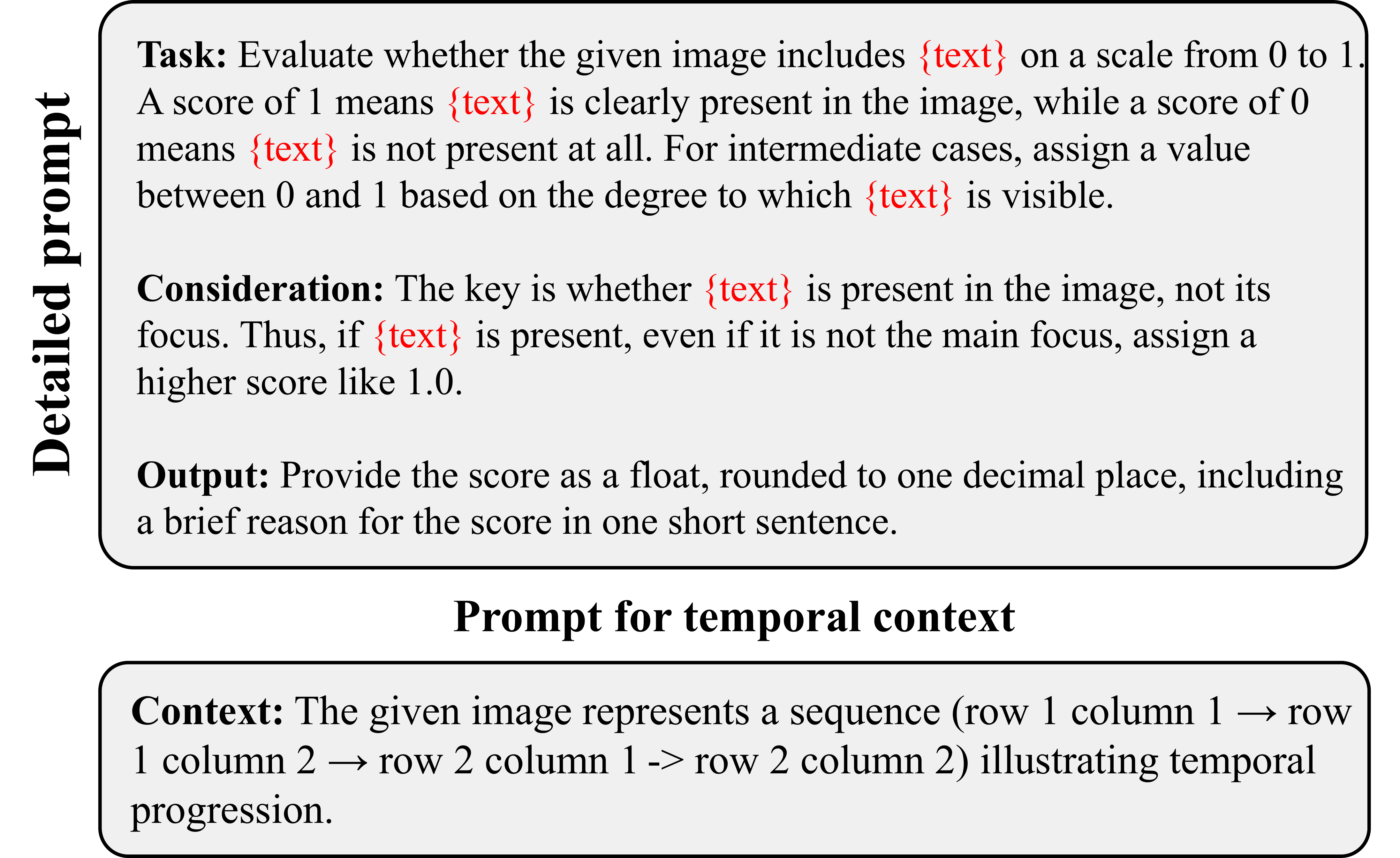}

   \caption{Proposed prompt for VQA}
   \label{fig4}
\end{figure}

\noindent\textbf{Prompt Design.}
\Cref{fig4} illustrates the proposed prompt $P$. The prompt comprises three main components: `task', `consideration', and `output’. First, `task’ defines the operation the LVLM should perform, specifically evaluating whether $X$ is present in the image. Next, `consideration’ specifies factors to be taken into account during evaluation, while `output’ defines the format for presenting the evaluation results. To leverage the chain-of-thought \cite{wei2022chain} effect, we instructed the model to provide brief reasoning along with the anomaly score, rounded to one decimal place. When conducting VQA using $TC$, an additional element, `context', is inserted between `task’ and `consideration’ in the prompt. This context element conveys the meaning of the rows and columns of $TC$ to the LVLM. We define the modified prompt as $P^*$. \\

\noindent\textbf{Anomaly Scoring.} 
The context serves as supplementary information to the image. However, because the LVLM accepts only a single image as input, it is challenging to utilize both the original and additional information simultaneously. To address this issue, we adopt a late fusion approach. Specifically, $\hat{k}$, $PC$, and $TC$ were used as the image inputs for the LVLM. The LVLM returns an anomaly score for each input, and these three scores are combined to compute the final $ascore$: 
\begin{equation}
\begin{aligned}
ascore = & \ \gamma_1 \cdot \Phi_{\text{LVLM}}(\hat{k}, P) + \gamma_2 \cdot \Phi_{\text{LVLM}}(PC, P) \\
& + \gamma_3 \cdot \Phi_{\text{LVLM}}(TC, P^*)
\end{aligned}
\end{equation}
Here, $\gamma$ is a hyperparameter that adjusts the proportion of context reflected in $ascore$. A performance comparison experiment based on the hyperparameter tuning is presented in supplementary materials.

Consequently, even if the abnormal frame $\hat{k}$ receives a low score, the final anomaly score will be high if the additional information, $PC$ or $TC$, is assigned a high score. This enabled accurate anomaly detection. Finally, to create frame-level anomaly scores, we duplicated the $ascore$ for the length of each segment and then applied a temporal 1-D Gaussian filter for smoothing, following prior works \cite{reissattribute, wang2022video}.

\section{Experiments}
\subsection{Datasets}
\begin{figure}[t]
  \centering
   \includegraphics[width=\linewidth]{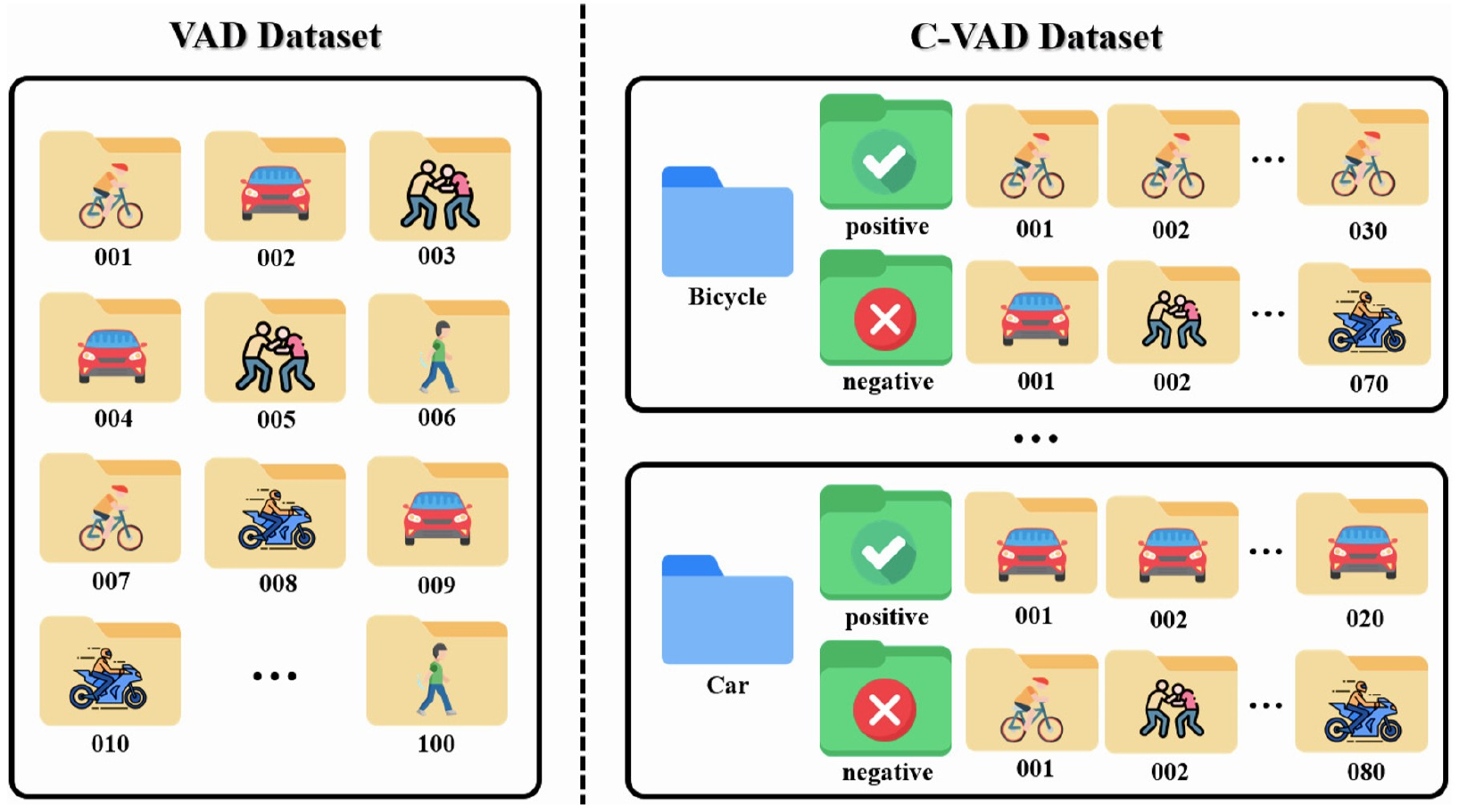}

   \caption{Comparison between the VAD and C-VAD datasets}
   \label{fig5}
\end{figure}

\Cref{fig5} illustrates the composition of the VAD and proposed C-VAD datasets. In conventional VAD datasets, videos are not categorized by an abnormal class type. In contrast, the proposed C-VAD datasets are organized by abnormal event type, with videos classified as positive or negative based on the presence of each abnormality. This categorization enables a precise evaluation of detection performance for specific types of abnormalities (e.g., bicycle). In this study, we validated the effectiveness of the proposed method on four VAD datasets: CUHK Avenue (Ave) \cite{lu2013abnormal}, ShanghaiTech Campus (ShT) \cite{luo2017revisit}, UBnormal (UB) \cite{acsintoae2022ubnormal}, and UCF-Crime (UCF) \cite{sultani2018real} as well as two C-VAD datasets: Customizable-ShT (C-ShT) and Customizable-Ave (C-Ave). Further details of the datasets are provided in supplementary materials.

\subsection{Evaluation Criteria}
To ensure consistency with previous VAD studies, the performance of the proposed model was evaluated using the micro-averaged area under the receiver operating characteristic curve (micro AUROC) metric. Specifically, the anomaly scores of all the frames in the dataset were aggregated, and the threshold of the anomaly score was progressively adjusted to compute the final evaluation.

\begin{table*}[t!]
\centering
\caption{Performance comparison on C-ShT dataset. The best
results are \textbf{bolded}. The second-best results are \underline{underlined}.}
\label{tab:comparison_methods}
\resizebox{\linewidth}{!}{
\begin{tabular}{llcccccc}
\toprule
\textbf{Category} & \textbf{Class} & \textbf{Baseline} & \textbf{+KSM} & \textbf{+KSM/PC} & \textbf{+KSM/TC} & \textbf{Proposed} & \textbf{Improvement (\%)} \\
\midrule
\multirow{7}{*}{\textbf{Action}} 
& Skateboarding & \underline{61.30} & 57.06 & 57.79 & \textbf{73.66} & \textbf{73.66} & +20.16 \\
& Throwing      & \textbf{91.41} & 72.82 & 88.74 & 82.53 & \underline{90.67} & -0.81 \\
& Running       & 53.13 & 51.93 & 53.68 & \underline{59.77} & \textbf{60.11} & +13.14 \\
& Loitering     & 61.98 & 51.96 & \textbf{81.27} & \underline{76.94} & \textbf{81.27} & +31.12 \\
& Jumping       & 82.84 & 92.89 & \underline{92.91} & \textbf{95.31} & \textbf{95.31} & +15.05 \\
& Falling       & 78.31 & 78.95 & 79.24 & \textbf{88.01} & \textbf{88.01} & +12.39 \\
& Fighting      & 84.48 & \underline{91.18} & \underline{91.18} & \textbf{98.06} & \textbf{98.06} & +16.07 \\
\cmidrule{2-8}
& \textbf{Average} & 73.35 & 72.00 & 77.83 & \underline{82.04} & \textbf{83.87} & +14.34 \\
\midrule
\multirow{5}{*}{\textbf{Appearance}} 
& Car           & 88.72 & \underline{90.96} & \textbf{91.46} & \underline{90.96} & \textbf{91.46} & +3.09 \\
& Hand truck    & 95.50 & 98.20 & \underline{98.91} & \textbf{99.20} & \textbf{99.20} & +3.87 \\
& Bicycle       & 72.36 & \underline{72.46} & \textbf{78.47} & \underline{72.46} & \textbf{78.47} & +8.44 \\
& Motorcycle    & \textbf{88.04} & \underline{86.72} & \underline{86.72} & \underline{86.72} & \underline{86.72} & -1.50 \\
\cmidrule{2-8}
& \textbf{Average} & 86.16 & 87.09 & \underline{88.89} & 87.34 & \textbf{88.95} & +3.25 \\
\midrule
\rowcolor{blue!10}
\textbf{Overall Average} & & 78.01 & 77.48 & 81.85 & \underline{83.97} & \textbf{85.72} & +9.88 \\
\bottomrule
\end{tabular}
}
\label{tab:tab1}
\end{table*}

\begin{table*}[t!]
\centering
\caption{Performance comparison on C-Ave dataset}
\label{tab:comparison_methods}
\resizebox{\linewidth}{!}{
\begin{tabular}{llcccccc}
\toprule
\textbf{Category} & \textbf{Class} & \textbf{Baseline} & \textbf{+KSM} & \textbf{+KSM/PC} & \textbf{+KSM/TC} & \textbf{Proposed} & \textbf{Improvement (\%)} \\
\midrule
\multirow[c]{4}{*}{\textbf{Action}} 
& Throwing & 78.44 & 80.13 & \textbf{89.77} & \underline{82.40} & \textbf{89.77} & +14.44 \\
& Running      & 75.82 & \underline{77.67} & \underline{77.67} & \textbf{77.90} & \textbf{77.90} & +2.74 \\
& Dancing       & \underline{85.65} & 72.28 & 76.64 & \textbf{91.92} & \textbf{91.92} & +7.32 \\
\cmidrule{2-8}
& \textbf{Average} & 79.97 & 76.69 & 81.36 & \underline{84.07} & \textbf{86.53} & +8.2 \\
\midrule
\multirow[c]{3}{*}{\textbf{Appearance}} 
& Too close           & 57.23 & \underline{61.48} & \underline{61.48} & \textbf{91.78} & \textbf{91.78} & +60.37 \\
& Bicycle    & \underline{99.99} & 99.84 & \underline{99.99} & 99.93 & \textbf{100.00} & +0.01 \\
\cmidrule{2-8}
& \textbf{Average} & 78.61 & 80.66 & 80.74 & \underline{95.86} & \textbf{95.89} & +21.98 \\
\midrule
\rowcolor{blue!10}
\textbf{Overall Average} & & 79.43 & 78.28 & 81.11 & \underline{88.79} & \textbf{90.27} & +13.65 \\
\bottomrule
\end{tabular}
}
\label{tab:tab2}
\end{table*}

\subsection{Results}
\Cref{tab:tab1,tab:tab2} present the evaluation results on the C-VAD datasets. The baseline, as described in \Cref{sec:intro}, performs VQA at the frame level to compute anomaly scores. The proposed model achieved performance improvements of 9.88\% and 13.65\% compared to the baseline on the C-ShT and C-Ave datasets, respectively. Specifically, it showed improvements of 14.34\% and 8.2\% in the action class, and 3.25\% and 21.98\% in the appearance class, respectively.

When only KSM was applied to the baseline, the execution time decreased in proportion to the segment length, whereas the average performance remained similar to that of the baseline. This is because the CLIP effectively selects representative frames for each segment, thereby compensating for the loss of temporal information. However, because it does not fully capture fine-grained spatio-temporal details, its performance significantly decreases for certain classes.  Therefore, we address these issues using the proposed contextual information. First, using $PC$ resulted in performance improvements of 5.64\% and 3.62\% compared to the KSM, as the LVLM focused on analyzing objects related to $X$. Additionally, applying $TC$ led to performance improvements of 8.38\% and 14.43\% over the KSM, respectively, with particularly notable enhancements observed in the action class. This indicates that utilizing the temporal information provided in the grid image is essential for action analysis. Additional validations, such as FPS comparisons by segment length and performance evaluations across various LVLMs, are presented in supplementary materials.

\subsection{Qualitative Analysis}
\begin{figure*}[!t]
    \centering
    \begin{subfigure}[h]{0.95\linewidth}
        \centering
        \includegraphics[width=\textwidth]{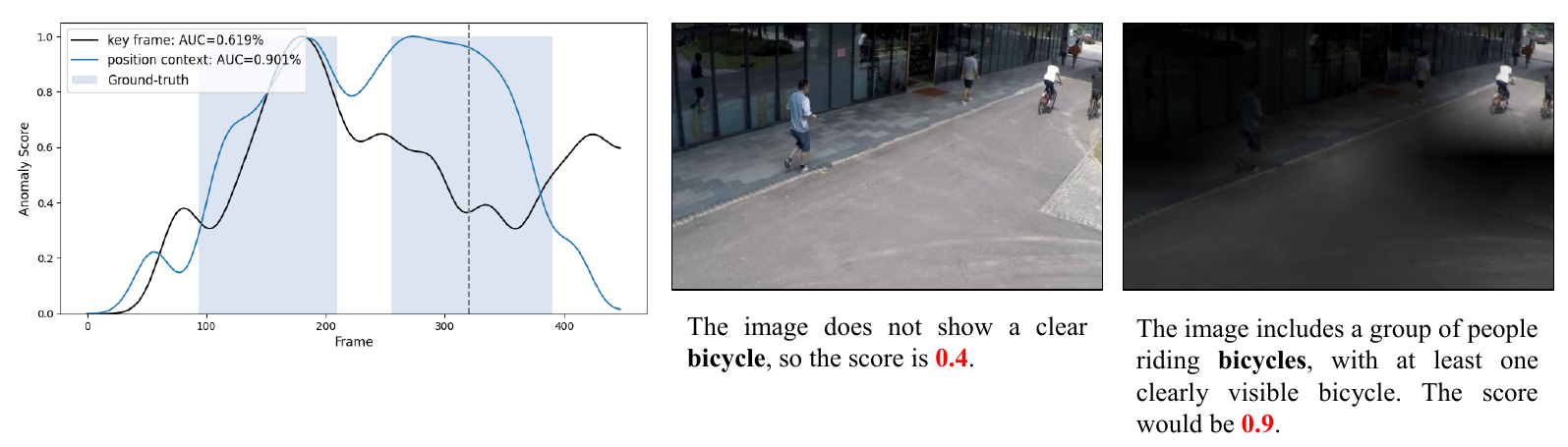}
    \end{subfigure}
    \begin{subfigure}[h]{0.95\linewidth}
        \centering
        \includegraphics[width=\textwidth]{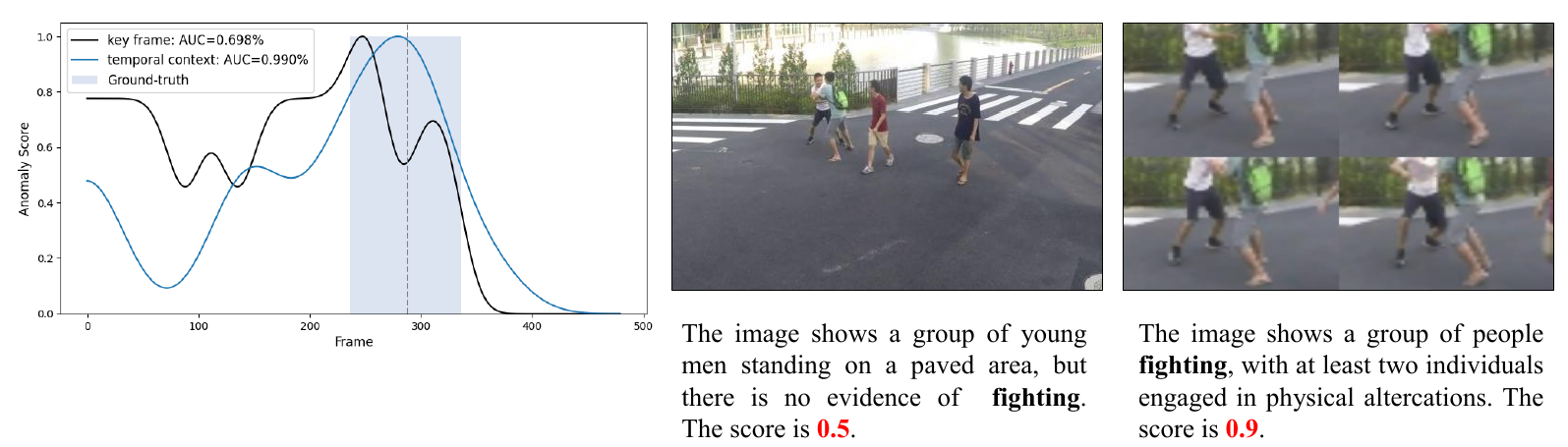}
    \end{subfigure}
    \caption{Anomaly score comparison with context visualization}
    \label{fig:figure6}
\end{figure*}

To analyze the effect of context-aware VQA, we present the visualization results for the anomaly scores and input frames in \Cref{fig:figure6}. When $PC$ was not applied, the bicycle object appeared smaller then the other objects, leading to a lower detection performance. Once $PC$ is applied, the bicycle region is emphasized, thereby enhancing the object recognition capability of the LVLM. Similarly, without $TC$, the model misinterpreted fighting as standing, resulting in lower detection performance. Incorporating temporal information through $TC$ improves the action recognition capability of the LVLM. These results demonstrate that context-aware VQA is more effective than the conventional VQA.

\subsection{Ablation Study}
\label{sec:ablation_study}

\begin{table}[h]
    \caption{Comparison on key frames selection method. RD, CP and Gr. indicate random, CLIP and grouping, respectively. * indicates testing without context. Act. and App. indicate action and appearance class, respectively.}
    \centering
    \begin{tabular}{ccccccc}
        \toprule
        \multirow{2}{*}{\shortstack{\textbf{Key} \\ \textbf{frames}}} & \multicolumn{3}{c}{\textbf{C-ShT}} & \multicolumn{3}{c}{\textbf{C-Ave}} \\
        \cmidrule(lr){2-4} \cmidrule(lr){5-7}
        & Act. & App. & Total & Act. & App. & Total \\
        \midrule
        \rowcolor{gray!10}
        RD* & 69.9 & 84.0 & 75.0 & 66.4 & 78.8 & 71.3 \\
        \rowcolor{gray!10}
        CP* & 72.0 & 87.1 & 77.5 & 76.7 & 80.7 & 78.3 \\
        \hline
        RD & 80.0 & \textbf{89.1} & 83.3 & 79.1 & 92.3 & 84.4 \\
        CP & 81.2 & 88.9 & 84.0 & 84.3 & 81.2 & 83.1 \\
        Gr. → CP & 82.2 & 88.8 & 84.7 & 83.9 & 92.2 & 87.2  \\
        \rowcolor{blue!10}
        CP → Gr. & \textbf{83.9} & 89.0 & \textbf{85.7} & \textbf{86.5} & \textbf{95.9} & \textbf{90.3} \\
        \bottomrule
    \end{tabular}
    \label{tab:tab3}
\end{table}

\noindent\textbf{Key frames Selection.}
We conducted an ablation study on key frames selection from two perspectives: temporal uniformity and text alignment. The random method considers neither of these aspects, whereas the CLIP-based approach considers only text alignment. Selecting key frames using CLIP after grouping ensures text alignment but does not guarantee temporal uniformity. Applying grouping after CLIP resulted in evenly distributed key frames, thereby considering both temporal uniformity and text alignment. As shown in \Cref{tab:tab3}, incorporating both factors yielded the best performance for C-VAD, highlighting the critical role of temporal uniformity in action recognition. Furthermore, RD* and CP*, which do not utilize the contextual information, perform worse than the random method, which disregards both temporal uniformity and text alignment. This demonstrates the importance of leveraging the contextual information.\\

\noindent\textbf{Window Size.} 
\Cref{tab:tab4} presents the experimental results based on the window sizes used in $PC$ and $TC$. For the action classes, the best performance was achieved with the large window size in C-ShT and the middle window size in C-Ave. This indicates that middle or large window sizes are more effective in capturing temporal movements and interactions between multiple objects. For appearance classes, the optimal performance was observed with the small window size in C-ShT and the middle window size in C-Ave, suggesting that the appropriate window size varies depending on the dataset owing to differences in object sizes. To enhance the generalization performance of the model, we adopted an approach that utilized all three window sizes and found that incorporating them yielded the best overall performance. Further details on the window size are provided in the implementation details section of the supplementary material.

\begin{table}[t]
    \caption{Comparison on window size.}
    \centering
    \begin{tabular}{ccccccc}
        \toprule
        \multirow{2}{*}{\shortstack{\textbf{Window} \\ \textbf{Size}}} & \multicolumn{3}{c}{\textbf{C-ShT}} & \multicolumn{3}{c}{\textbf{C-Ave}} \\
        \cmidrule(lr){2-4} \cmidrule(lr){5-7}
        & Act. & App. & Total & Act. & App. & Total \\
        \midrule
        small & 78.8 & \textbf{90.6} & 83.1 & 84.7 & 87.1 & 85.7 \\
        middle & 81.2 & 89.0 & 84.1 & \textbf{87.5} & 92.0 & 89.3 \\
        large & 82.1 & 89.7 & 84.9 & 86.8 & 86.4 & 86.6  \\
        \rowcolor{blue!10}
        all & \textbf{83.9} & 89.0 & \textbf{85.7} & 86.5 & \textbf{95.9} & \textbf{90.3} \\
        \bottomrule
    \end{tabular}
    \label{tab:tab4}
\end{table}

\begin{table}[t]
\caption{Comparison with state-of-the-art VAD methods. * indicates testing without context.}
\centering
\resizebox{\linewidth}{!}{
\begin{tabular}{cccccc}
\toprule
\textbf{Method} & \textbf{Zero-shot} & \textbf{Ave} & \textbf{ShT} & \textbf{UB} & \textbf{UCF}\\
\addlinespace[2pt]
\hline
AMMC-Net\cite{cai2021appearance} & \ding{55} & 86.6 & 73.7 & - & - \\
STEAL-Net\cite{astrid2021synthetic} &  \ding{55} & 87.1 & 73.7 & - & - \\
MPN\cite{lv2021learning} & \ding{55} & 89.5 & 73.8 & - & - \\
DLAN-AC\cite{yang2022dynamic} &  \ding{55} & 89.9 & 74.7 & - & - \\
UBnormal\cite{acsintoae2022ubnormal} & \ding{55} & - & - & 68.5 & - \\
FPDM\cite{yan2023feature} & \ding{55} & 90.1 & 78.6 & 62.7 & 74.7 \\
SLM\cite{shi2023video} & \ding{55} & \underline{90.9} & 78.8 & - & - \\
USTN-DSC\cite{yang2023video} & \ding{55} & 89.9 & 73.8 & - & - \\
AnomalyRuler\cite{Yang2024FollowTR} & \ding{55} & 89.7 & \textbf{85.2} & 71.9 & - \\
MULDE\cite{micorek2024mulde} &  \ding{55} & - & \underline{81.3} & 72.8 & \underline{78.5} \\
AED-MAE\cite{ristea2024self} & \ding{55} & \textbf{91.3} & 79.1 & 58.5 & - \\
MA-PDM\cite{zhou2024video} & \ding{55} & \textbf{91.3} & 79.2 & 63.4 & - \\
AccI-VAD\cite{reissattribute} &  \ding{55} & - & 76.2 & 66.8 & 60.3 \\
\hline
\rowcolor{blue!10}
AnyAnomaly* & \ding{51} & 81.4 & 77.2 & \underline{73.1} & 77.8\\
\rowcolor{blue!10}
AnyAnomaly & \ding{51} & 87.3 & 79.7 & \textbf{74.5} & \textbf{80.7}\\
\bottomrule
\end{tabular}
}
\label{tab:tab5}
\end{table}

\begin{table}[ht]
\caption{Generalization performance comparison. Tr.: cross-domain training where models trained on one VAD dataset are evaluated on another. Few.: methods that adapt to the target domain using only a few training samples,  Aux.: methods that utilize auxiliary datasets, *: since competitors did not perform cross-domain evaluations on ShT, we present their same-domain results instead.}
\centering
{
\begin{tabular}{cccccc}
\toprule
\textbf{Method} & \textbf{Tr.} & \textbf{Few.} & \textbf{Aux.} & \textbf{Ave} & \textbf{ShT}\\
\addlinespace[2pt]
\hline
STEAL-Net\cite{astrid2021synthetic} & \ding{51} & \ding{55} & \ding{55} & 54.3 & 51.7 \\ 
Jigsaw\cite{wang2022video} & \ding{51} & \ding{55} & \ding{55} & 62.9 & 59.3 \\ 
\hline
\rowcolor{gray!10}
rGAN\cite{lu2020few} & \ding{51} & \ding{51} & \ding{55} & 76.6 & 77.9* \\ 
\rowcolor{gray!10}
MPN\cite{lv2021learning} & \ding{51} & \ding{51} & \ding{55} & 78.9 & 73.8* \\ 
\rowcolor{gray!10}
zxVAD\cite{aich2023cross} & \ding{51} & \ding{55} & \ding{51} & 82.2 & 71.6* \\ 
\rowcolor{gray!10}
Shibao et al.\cite{gao2024scene} & \ding{51} & \ding{55} & \ding{51} & \underline{86.2} & \underline{78.7} \\ 
\hline
\rowcolor{blue!10}
ZS CLIP\cite{radford2021learning} & \ding{55} & \ding{55} & \ding{55} & 62.3 & 60.9 \\ 
\rowcolor{blue!10}
ZS ImageBind\cite{girdhar2023imagebind} & \ding{55} & \ding{55} & \ding{55} & 64.5 & 61.3 \\ 
\rowcolor{blue!10}
LLaVA-1.5\cite{liu2024improved} & \ding{55} & \ding{55} & \ding{55} & 67.4 & 59.6 \\ 
\rowcolor{blue!10}
Video-ChatGPT\cite{maaz2024video} & \ding{55} & \ding{55} & \ding{55} & 76.9 & 69.1 \\ 
\rowcolor{blue!10}
AnyAnomaly & \ding{55} & \ding{55} & \ding{55} & \textbf{87.3} & \textbf{79.7} \\ 
\bottomrule
\end{tabular}
}
\label{tab:tab6}
\end{table}

\subsection{Comparison with SOTA}
\label{sec: comparison with SOTA}
To assess the effectiveness of AnyAnomaly in handling multiple text inputs, we conducted experiments on the VAD benchmark datasets. For performance evaluation, each anomaly class in the dataset was treated as $X$, and the maximum anomaly score among all computed scores was assigned to the corresponding segment. \Cref{tab:tab5} presents a performance comparison with frame-centric VAD methods. Despite not being trained on VAD datasets, AnyAnomaly demonstrated a performance comparable to that of SOTA methods. Notably, it achieved the best AUC results on the UB and UCF datasets, which cover diverse scenes and abnormal events, thereby demonstrating the effectiveness of the proposed model across various environments. Furthermore, while LLM-based methods (e.g., AnomalyRuler) require rule generation and aggregation using a few normal samples, the proposed method achieves competitive performance solely through zero-shot inference, highlighting its practical applicability.

\subsection{Generalization Performance Comparison}
\Cref{tab:tab6} presents a comparison of the generalization performance of AnyAnomaly. Although STEAL-Net \cite{astrid2021synthetic} and Jigsaw \cite{wang2022video} achieved high accuracy in same-domain testing, their performance was significantly degraded in cross-domain settings. Specifically, on the Ave dataset, the performances of STEAL-Net and Jigsaw decreased as 87.1\% → 54.3\% and 92.2\% → 62.9\%, respectively. Similarly, on the ShT dataset, their performance decreased as 73.7\% → 51.7\% and 84.3\% → 59.3\%, respectively. This suggests that the existing OCC-based VAD models tend to overfit the training data, making them less effective when applied to new environments. For instance, `Too close' where an object is in close proximity to the camera is considered anomalous in the Ave dataset but normal in the ShT dataset. Consequently, OCC-based models trained on ShT struggle to detect such anomalies.

The zero- and few-shot VAD models designed for xVAD exhibited better generalization performance than the OCC-based models. However, few-shot models depend heavily on the number of K-shot samples, whereas zero-shot models require auxiliary datasets. Training-free methods using VLMs, such as ZS-CLIP and Video-ChatGPT, leverage strong image and video understanding capabilities, outperforming some VAD models. Nevertheless, their performance is limited by domain gaps. In contrast, AnyAnomaly effectively overcomes these gaps by incorporating contextual information, achieving superior performance.
\section{Conclusion}
We propose AnyAnomaly, a novel approach that leverages the LVLM for universal VAD. AnyAnomaly effectively performs the C-VAD by incorporating a segment-level approach and context-aware VQA. This design reduces latency when processing large videos and minimizes the domain gap between the LVLM and VAD task. Despite being a zero-shot method, AnyAnomaly demonstrates competitive performance on benchmark datasets and holds promise for real-world VAD. Furthermore, because it operates without any training and enables anomaly detection in any video, it significantly improves accessibility in the VAD domain. We anticipate that AnyAnomaly will contribute substantially to VAD research and practical deployment.

\section*{Acknowledgements}

This work was supported by the National Research Foundation of Korea(NRF) grant funded by the Korea government(MSIT).(No. RS-2025-02312833)

{
    \small
    \bibliographystyle{ieeenat_fullname}
    \bibliography{main}
}
\clearpage
\setcounter{section}{0}
\renewcommand\thesection{\Alph{section}}
\setcounter{figure}{0}
\renewcommand\thefigure{S\arabic{figure}}
\setcounter{table}{0}
\renewcommand\thetable{S\arabic{table}}

\appendix

\maketitlesupplementary

% \noindent In this supplement, we provide the followings:
% \begin{itemize}
% \item \ref{supply1}: Additional quantitative evaluation for anomaly score hyperparameter and Large Vision Language Models.
% \item \ref{supply2}: Additional qualitative evaluation for context complemen
% \end{itemize}

\section{Experiment Details}
\label{supply1}

\subsection{Dataset Details}
\label{supply1.1}
\noindent\textbf{VAD Dataset.}
We used the CUHK Avenue (Ave) \cite{lu2013abnormal}, ShanghaiTech Campus (ShT) \cite{luo2017revisit}, UBnormal (UB) \cite{acsintoae2022ubnormal}, and UCF-Crime (UCF) \cite{sultani2018real} datasets. Ave comprises of videos captured by a single camera on a university campus, containing five types of abnormal events; throwing paper, running, dancing, approaching the camera (Too close) and bicycle. ShT is a campus CCTV dataset that includes 13 different background scenes and 11 types of abnormal events; such as bicycles, cars, fighting, and jumping. UB is a synthetic dataset generated using the Cinema4D software, encompassing 29 diverse background scenes, including indoor environments, sidewalks, and \etc. It provides a total of 22 abnormal events, including not only challenging-to-detect events such as smoking and stealing but also complex scenarios such as driving outside the lane and people-car accidents. UCF is a large-scale dataset containing 1,900 untrimmed real-world surveillance videos, covering 13 anomaly categories such as assault, burglary, explosion, and stealing.\\

\noindent\textbf{C-VAD Dataset.}
We constructed the Customizable-ShT (C-ShT) and Customizable-Ave (C-Ave) datasets. C-ShT reorganizes the test data of ShT into 11 abnormal event types and assigns new labels to each type. For example, in the bicycle category, videos containing bicycles were assigned to positive, whereas all other videos were assigned to negative. The frame-level labels were set to 1 only for frames in which a bicycle appeared in the positive videos. C-Ave was constructed by reorganizing the test data of Ave into 5 abnormal event types, following the same labeling methodology as C-ShT.

\subsection{Implementation Details}
\label{supply1.2}
In a key experiment using the C-VAD datasets, we employed an efficient Chat-UniVi \cite{jin2024chat} 7B model, considering the balance between performance and speed. For the VAD dataset experiment, we utilized the effective MiniCPM-V \cite{yao2024minicpm} 8B model to achieve optimal performance and compared it with state-of-the-art (SOTA) models. At the time, both models represented a well-validated choice for context-aware VQA. Additional results with Qwen2.5-VL \cite{Qwen2.5-VL} are reported in \Cref{tab:sup4} for further comparison. The CLIP model used for key frames selection and context generation was ViT-B/32. For context generation, we adopted large, middle, and small window sizes of (120,120), (80,80), and (48,48), respectively. For C-Ave and Ave, the large window size was set to (240,240). All the experiments were conducted on a single NVIDIA GeForce RTX 3090 GPU.

\subsection{Prompt Details}
\label{supply1.3}

\begin{figure*}[t]
  \centering
   \includegraphics[width=0.85\linewidth]{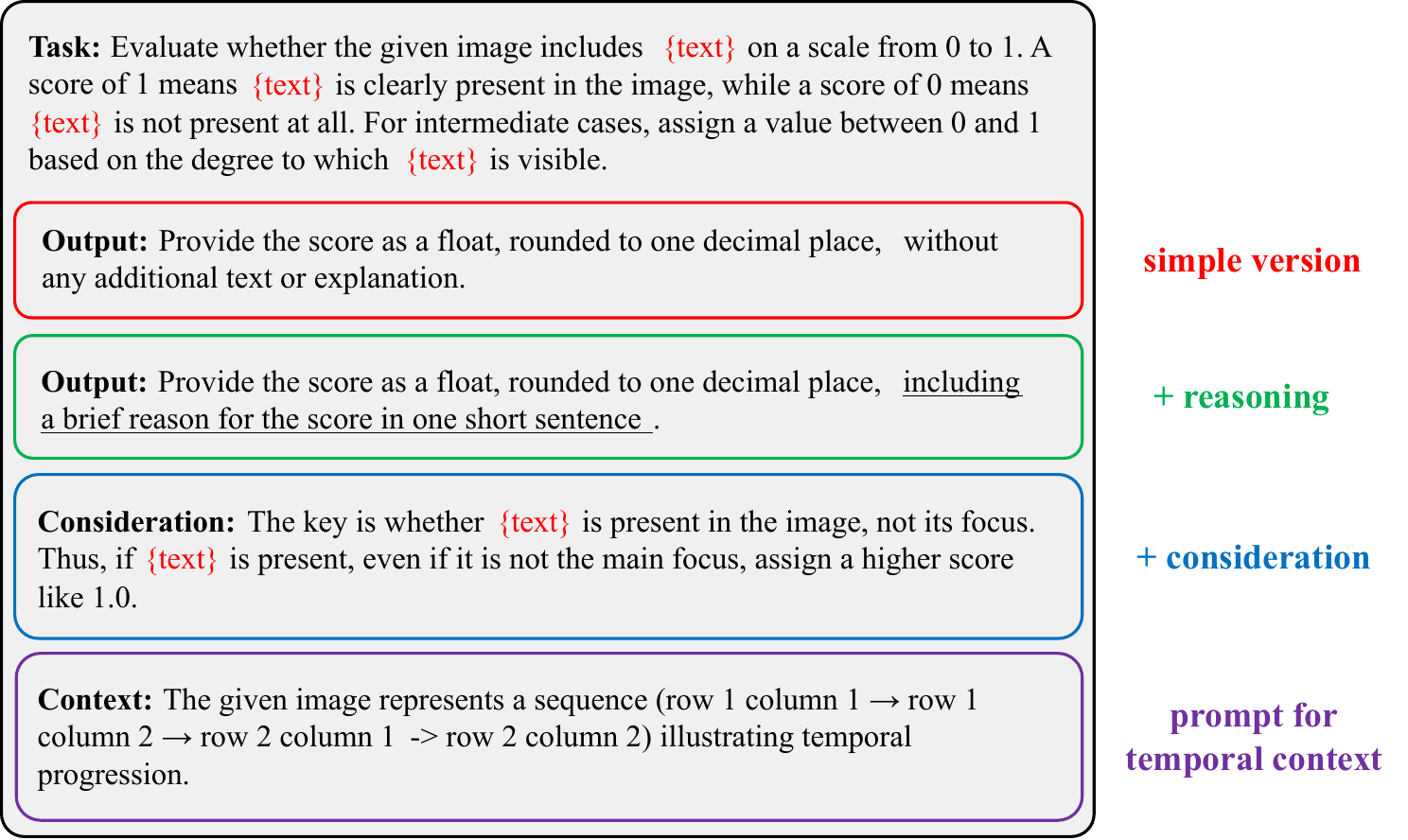}
   \caption{Prompt details. The content written in the simple version is not utilized when applying reasoning.}
   \label{fig:figureS1} 
\end{figure*}

% \section{Prompt details}
% \label{supply1}
\Cref{fig:figureS1} shows the detailed prompts used in the experiments. First, a reasoning prompt is designed to obtain the chain-of-thought \cite{wei2022chain} effect by requiring a simple reason along with the anomaly score. This helps to break down the problem step-by-step, guiding the model to resolve complex issues more systematically. For example, the question \textit{“Does the image include jumping?} can be divided into two steps: \textit{1. “Is there an object related to jumping (e.g., a person)?”} and \textit{2. “Is the object performing a jumping action?”} This allows object-level image analysis, leading to more refined predictions. The consideration prompt encourages the assignment of a high score even when $X$ is not central within the image. This prompt was introduced to address the issue where low scores are assigned simply because $X$ exists but is not the central element. The effectiveness of this prompt tuning is compared and analyzed in \Cref{tab:sup1}.

The simple prompt instructs the LVLM to output only the anomaly score, while adding reasoning prompt the model to perform reasoning during the score calculation process, and applying consideration prompt encourages the model to focus on the given text. Experimental results showed that using both reasoning and consideration prompt achieved the best performance, suggesting that when the LVLM includes reasoning in the process, it produces more accurate results and can respond more precisely to user instructions through consideration prompt.

\begin{table}[t]
\caption{Comparison on prompt tuning}
\centering
\begin{tabular}{ccc}
\toprule
\textbf{Prompt Tuning} & \textbf{C-ShT} & \textbf{C-Ave} \\
\addlinespace[2pt]
\hline
Baseline (simple) & 70.38 & 67.58 \\ 
Baseline (+reasoning) & 71.58 & 72.79\\ 
\rowcolor{gray!10}
Baseline (+reasoning, consideration) & 78.01 & 79.43 \\ 
\hline
Proposed (simple) & 79.29 & 74.01 \\ 
Proposed (+reasoning) & 79.79 & 82.09\\ 
\rowcolor{blue!10}
Proposed (+reasoning, consideration) & \textbf{85.72} & \textbf{90.27} \\ 
\bottomrule
\end{tabular}
\label{tab:sup1}
\end{table}

\begin{table}[t]
\caption{Comparison on segment length}
\centering
{
\begin{tabular}{cccccc}
\toprule
\textbf{Segment length} & \textbf{C-ShT} & \textbf{C-Ave} & \textbf{FPS}\\
\addlinespace[2pt]
\hline
\rowcolor{gray!10}
Baseline & 78.01 & 79.43 & 0.96 \\ 
\hline
8 & 83.83 & 83.96 & 2.67 \\ 
16 & 83.45 & 87.45 & 4.49 \\ 
\rowcolor{blue!10}
24 & \textbf{85.72} & \textbf{90.27} & 6.67 \\ 
32 & 82.50 & 85.94 & \textbf{8.45} \\ 
\bottomrule
\end{tabular}
}
\label{tab:sup2}
\end{table}

\begin{table}[t]
    \centering
    \caption{Comparison of different methods on various datasets}
    \setlength{\tabcolsep}{10pt}
    \begin{tabular}{c l c c}
        \toprule
        \textbf{Dataset} & \textbf{Method} & \textbf{Value} & \textbf{AUC} \\
        \midrule
        \multirow{3}{*}{Ave} 
            &w/o context& - & 81.4 \\
            &\cellcolor{gray!10}w/o tuning&\cellcolor{gray!10} 1.0, 1.0, 1.0&\cellcolor{gray!10} \underline{84.4} \\
            &\cellcolor{blue!10}w/ tuning&\cellcolor{blue!10} 0.6, 0.3, 0.1 &\cellcolor{blue!10} \textbf{87.3} \\
        \midrule
        \multirow{3}{*}{ShT} 
            &w/o context& - & 77.2 \\
            &\cellcolor{gray!10}w/o tuning&\cellcolor{gray!10} 1.0, 1.0, 1.0 &\cellcolor{gray!10} \underline{79.4} \\
            &\cellcolor{blue!10}w/ tuning&\cellcolor{blue!10} 0.5, 0.3, 0.2 &\cellcolor{blue!10} \textbf{79.7} \\
        \midrule
        \multirow{3}{*}{UB} 
            &w/o context& - & 73.1 \\
            &\cellcolor{gray!10}w/o tuning&\cellcolor{gray!10} 1.0, 1.0, 1.0 &\cellcolor{gray!10} \underline{73.8} \\
            &\cellcolor{blue!10}w/ tuning&\cellcolor{blue!10} 0.6, 0.1, 0.3 &\cellcolor{blue!10} \textbf{74.5} \\
        \midrule
        \multirow{3}{*}{UCF} 
            &w/o context& - & 77.8 \\
            &\cellcolor{gray!10}w/o tuning&\cellcolor{gray!10} 1.0, 1.0, 1.0 &\cellcolor{gray!10} \underline{80.5} \\
            &\cellcolor{blue!10}w/ tuning&\cellcolor{blue!10} 0.6, 0.1, 0.3 &\cellcolor{blue!10} \textbf{80.7} \\
        \bottomrule
    \end{tabular}
    \label{tab:sup3}
\end{table}

\begin{table*}[ht]
    \centering
    \caption{Comparison of diverse LVLMs. The model highlighted in blue represents the most efficient model for the C-VAD task, while the one highlighted in purple indicates the most effective model. For further comparison, additional experiments were conducted using Qwen-based models. *: Experiment conducted using vLLM.}
    \begin{tabular}{ccccccc}
        \toprule
        \multirow{2}{*}{\textbf{LVLM}} & \multirow{2}{*}{\textbf{Pre-trained}} & \multicolumn{2}{c}{\textbf{C-ShT}} & \multicolumn{2}{c}{\textbf{C-Ave}} &  \multirow{2}{*}{\textbf{FPS}}\\
        \cmidrule(lr){3-4} \cmidrule(lr){5-6}
        & & w/o context & Proposed & w/o context & Proposed & \\
        \midrule
        \rowcolor{blue!10}
        Chat-UniVi\cite{jin2024chat} & Chat-UniVi-7B & 77.5 & \underline{85.7} & 78.3 & \underline{90.3} & 6.67 \\
        MiniGPT-4\cite{zhu2023minigpt} & LLaMA-2 Chat 7B & 54.0 & 67.4 & 53.9 & 55.3 & 1.26\\
        \rowcolor{violet!10}
        MiniCPM-V\cite{yao2024minicpm} & MiniCPM-Llama3-V-2\_5 (8B) & 87.7 & \textbf{90.1} & 86.3 & \textbf{91.0} & 1.36\\
        LLAVA++\cite{hanoona2024LLaVA++} & LLaVA-Meta-Llama-3-8B-Instruct-FT & 73.3 & 82.8 & 59.0 & 69.4 & 7.25\\
        \midrule
        Qwen2.5-VL\cite{Qwen2.5-VL} & Qwen2.5-VL-3B-Instruct & 89.0 & \underline{90.2} & 78.0 & 87.0 & 11.18\\
        Qwen2.5-VL*\cite{Qwen2.5-VL} & Qwen2.5-VL-3B-Instruct & 88.6 & \underline{90.2} & 78.3 & \underline{88.1} & 34.78\\
        Qwen2.5-VL*\cite{Qwen2.5-VL} & Qwen2.5-VL-7B-Instruct & 93.0 & \textbf{95.5} & 86.9 & \textbf{92.4} & 24.08\\
        \bottomrule
    \end{tabular}
    \label{tab:sup4}
\end{table*}

\section{Additional Quantitative Evaluation}
\label{supply2}

\subsection{Segment length and FPS}
\label{supply2.1}
\Cref{tab:sup2} presents the performance comparison and FPS based on different segment lengths. The baseline segment length was set to 1. It was observed that deriving anomaly scores at the segment level yields superior performance compared to the baseline, which relies on a single frame. The highest AUC performance was achieved when the segment length is set to 24, reaching 85.72\% and 90.27\% for C-ShT and C-Ave, respectively. However, excessively long segment length introduces irrelevant information into the temporal context, leading to a decrease in accuracy. Furthermore, performing VAD at the segment level resulted in a 594\% improvement in the FPS compared with the baseline.

\subsection{Hyperparameter Tuning}
\label{supply2.2}
We tuned the three hyperparameters $\gamma_1, \gamma_2$, and $\gamma_3$ used for the final anomaly score calculation for each VAD dataset. Each hyperparameter controls the influence of the anomaly score derived from the frame, position, and temporal contexts. As shown in \Cref{tab:sup3}, the optimal hyperparameter values vary across datasets owing to differences in object sizes and abnormal events. Additionally, comparing w/o context, which does not utilize context information, and w/o tuning, where all hyperparameters were set to the same value, we observed performance improvements of 3.0\%, 2.2\%, 0.7\%, and 2.7\%, even without hyperparameter tuning. In contrast, the performance differences owing to hyperparameter tuning were 2.9\%, 0.3\%, 0.7\% and 0.2\%, respectively. This demonstrates the effectiveness of our proposed approach in utilizing context information in VAD and proves that it achieves a strong generalization performance even without hyperparameter tuning.

\subsection{Diverse LVLM Comparison}
\label{supply2.3}
\Cref{tab:sup4} presents the results for C-ShT and C-Ave when using various LVLMs. We evaluated the performances of four SOTA LVLMs: Chat-UniVi \cite{jin2024chat}, MiniGPT-4 \cite{zhu2023minigpt}, MiniCPM-V \cite{yao2024minicpm}, and LLAVA++ \cite{hanoona2024LLaVA++}. All experiments were conducted using the default settings, and `Pre-trained' refers to the names of the pre-trained model weights. The experimental results demonstrate that incorporating the proposed context-aware VQA improves the performance of all LVLMs. Specifically, the use context-aware VQA leads to improvements ranging from 2.6\% to 24.8\%. Notably, even MiniCPM, which achieved the best performance without context-aware VQA, and showed additional improvements of 2.7\% and 5.4\% for C-ShT and C-Ave, respectively, when context-aware VQA was applied. This confirms that leveraging the proposed context-aware VQA is effective for C-VAD. Additionally, we observed that Chat-UniVi, with an FPS of 6.67, was the most efficient model, whereas MiniCPM-V achieved the highest performance on both datasets, scoring 90.1\% and 91.0\%, respectively. Therefore, as mentioned in \Cref{supply1.1}, Chat-UniVi was used for the C-VAD experiments and MiniCPM-V was used for the VAD dataset experiments.

\subsection{Additional Experiments with vLLM}
\label{supply2.4}
With recent advances in LLM inference libraries such as vLLM \cite{kwon2023efficient}, latency issues can be largely mitigated. vLLM supports continuous batching and KV-caching, which substantially improve inference speed. Our method is not restricted to a specific LVLM and can be applied to various models supported by vLLM.

To evaluate efficiency, we conducted experiments by batching the key frame, $PC$, and $TC$ during C-VAD. The experiments were conducted using Qwen2.5-VL \cite{Qwen2.5-VL}, one of the representative models supported by vLLM. The results, presented in the lower part of \Cref{tab:sup4}, show that using vLLM maintained the AUC while achieving a 211\% improvement in FPS (from 11.18 to 34.78). Furthermore, even with a 7B model, higher FPS was observed compared to other models. These findings suggest that AnyAnomaly has strong potential for real-time applications.

\section{Additional Qualitative Evaluation}
\label{supply3}

\subsection{Further Analysis on Contexts}
\label{supply3.1}
To provide a deeper understanding of the role of $PC$ and $TC$, we present both qualitative analyses of their effectiveness and representative failure cases.

\begin{figure*}[h]
    \centering
    \begin{subfigure}[h]{0.9\linewidth}
        \centering
        \includegraphics[width=\textwidth]{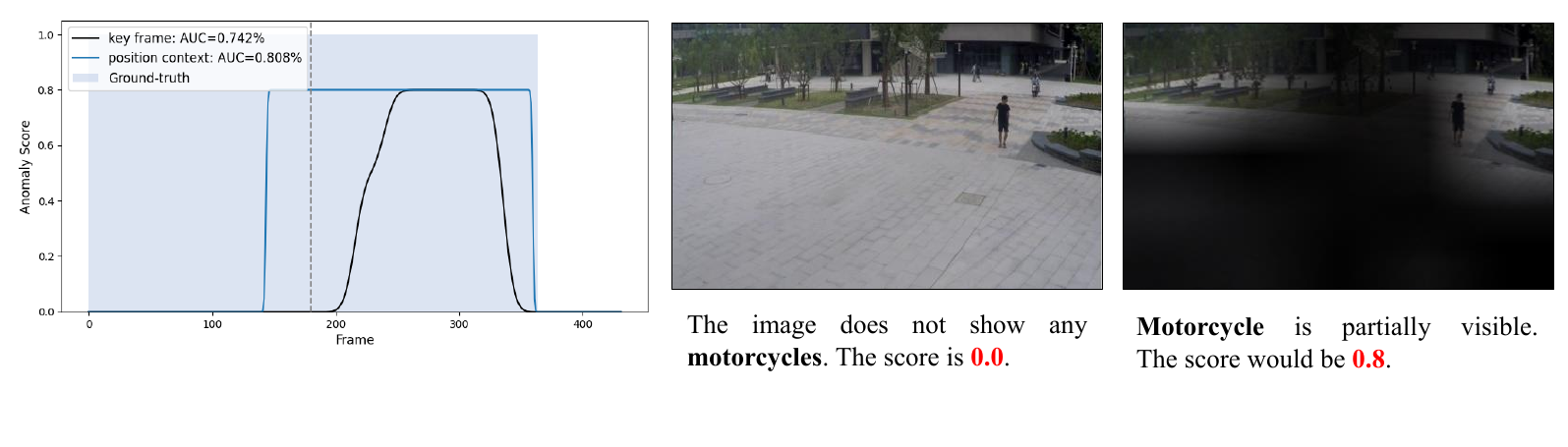}
    \end{subfigure}
    \begin{subfigure}[h]{0.9\linewidth}
        \includegraphics[width=\textwidth]{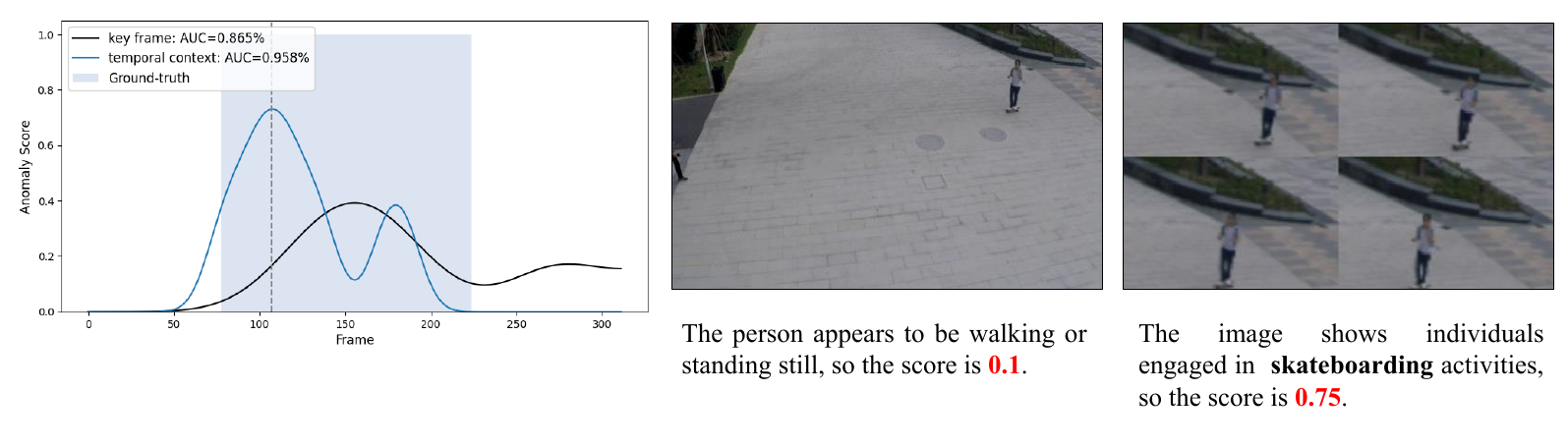}
    \end{subfigure}
    \caption{Anomaly score comparison with context visualization (success cases)}
    \label{fig:figureS2}
\end{figure*}

\Cref{fig:figureS2} illustrates the improvements achieved by incorporating $PC$ and $TC$. Without $PC$, the motorcycle appeared too small to be detected, resulting in an anomaly score of 0. With $PC$ applied, however, the relevant region was emphasized, and the score increased to 0.8 despite the object’s small size (top row). Likewise, without $TC$, the model misinterpreted skateboarding as walking and assigned a low anomaly score of 0.1. By integrating temporal information, the model captured motion cues such as positional changes and the enlarged appearance of the skateboard, correctly identifying skateboarding with a score of 0.75 (bottom row). These results indicate that our context-aware VQA is more effective than conventional VQA.

\begin{figure*}[h]
    \centering
    \begin{subfigure}[h]{0.9\linewidth}
        \centering
        \includegraphics[width=\textwidth]{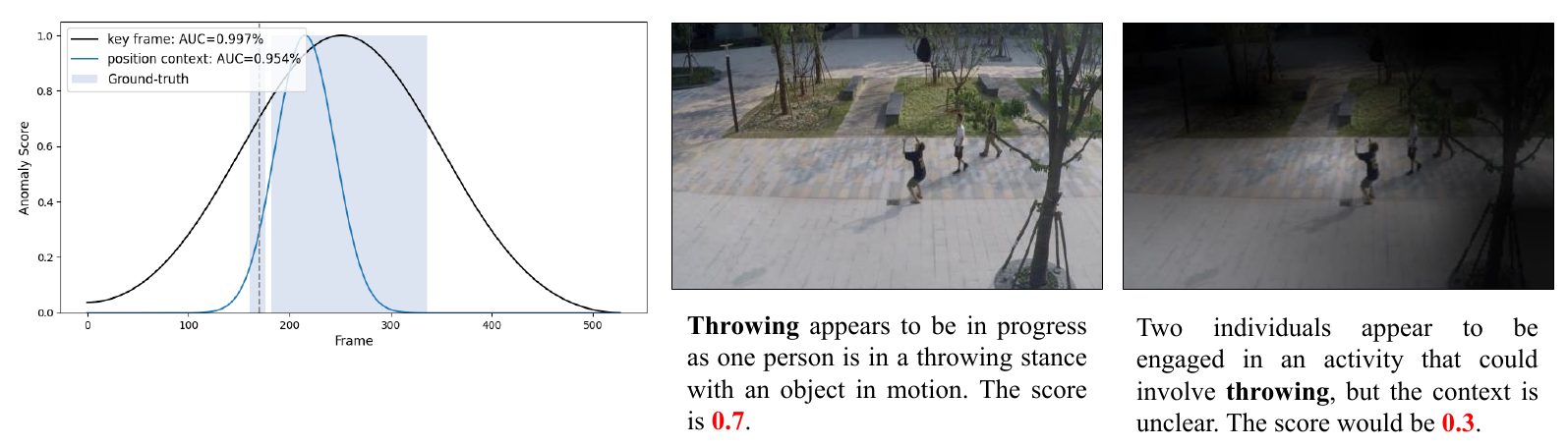}
    \end{subfigure}
    \begin{subfigure}[h]{0.9\linewidth}
        \includegraphics[width=\textwidth]{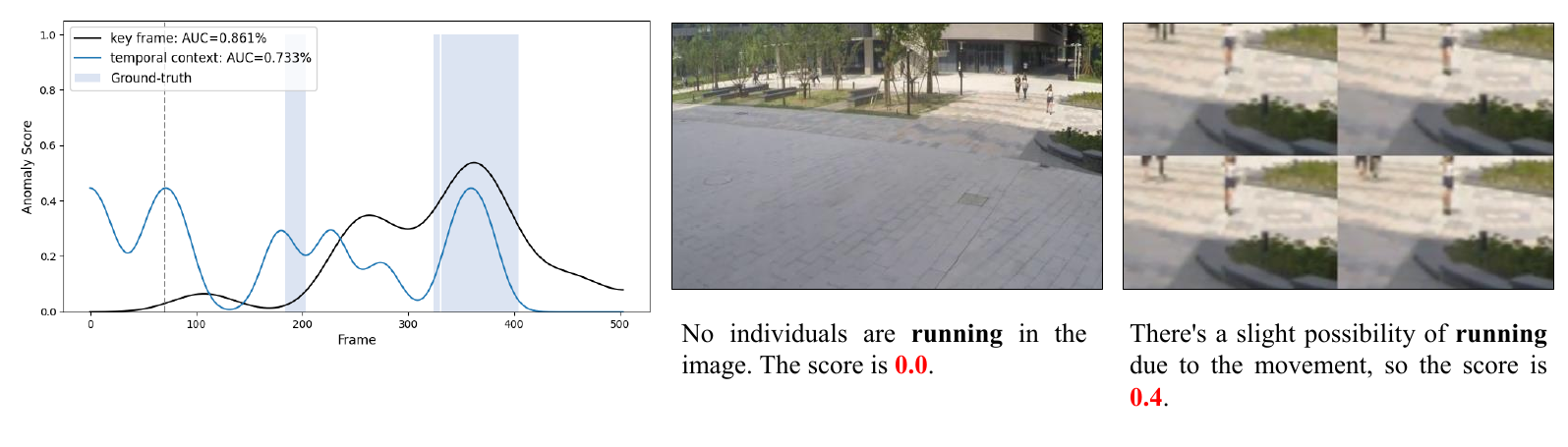}
    \end{subfigure}
    \caption{Anomaly score comparison with context visualization (failure cases)}
    \label{fig:figureS3}
\end{figure*}

Nevertheless, \Cref{fig:figureS3} demonstrates that $PC$ and $TC$ can also introduce errors under certain conditions. In the top row, before applying $PC$, the LVLM detected the abnormal action of throwing a bag with an anomaly score of 0.7. After applying $PC$, however, the model attended more to the person than the bag, making the action ambiguous and reducing the score to 0.3. In the bottom row, without $TC$, a normal walking scene yielded a score of 0, but after applying TC, walking was misinterpreted as a hard negative, raising the score to 0.4. While these failure cases exist, they provide valuable insights into the challenges of context-aware VQA and guide future directions for improvement.

\begin{figure*}[h]
    \centering
    \begin{subfigure}[h]{0.9\linewidth}
        \centering
        \includegraphics[width=\textwidth]{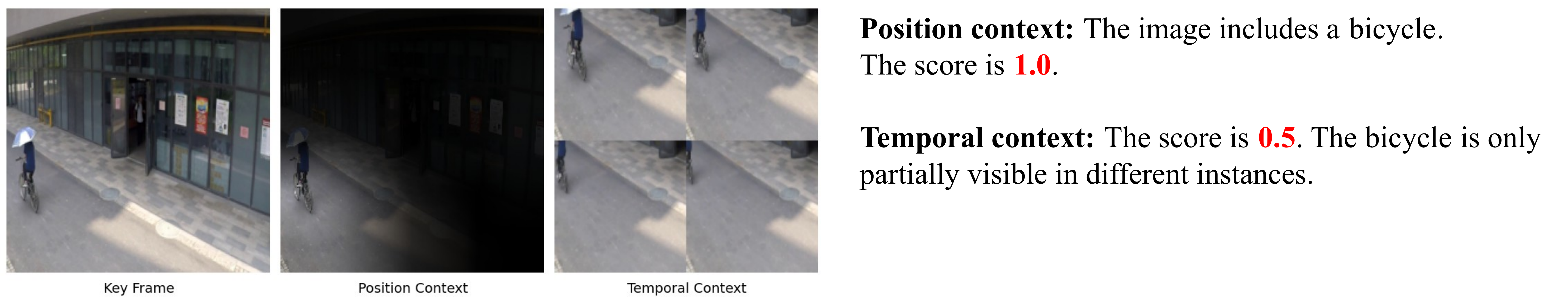}
    \end{subfigure}
    \begin{subfigure}[h]{0.9\linewidth}
        \includegraphics[width=\textwidth]{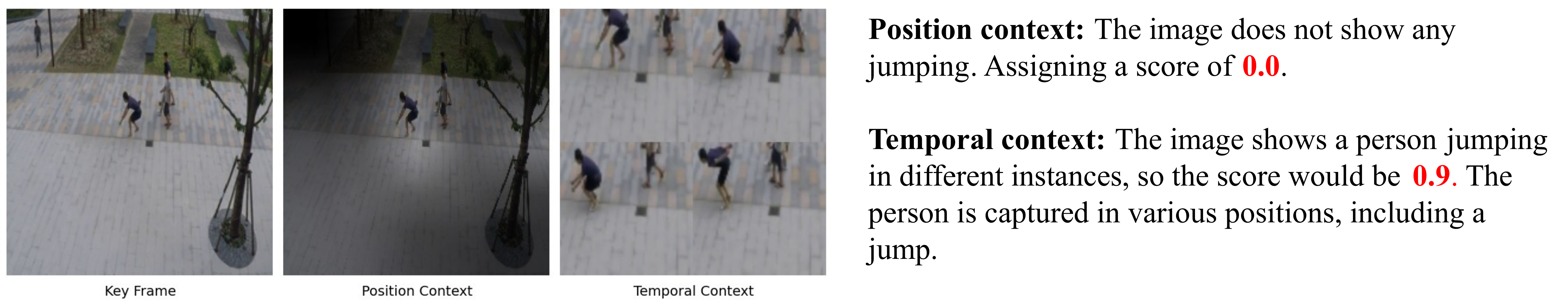}
    \end{subfigure}
    \caption{Example of complementarity between position and temporal context. The first example highlights the importance of position context and the second example emphasizes the importance of temporal context.}
    \label{fig:figureS4}
\end{figure*}

\begin{figure*}
    \centering
    \begin{subfigure}[h]{0.32\textwidth}
        \centering
        \includegraphics[width=\textwidth]{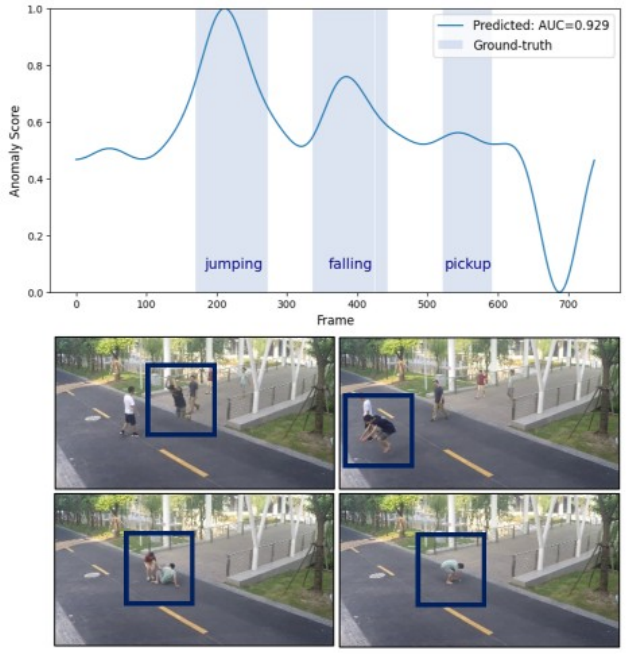}
        \caption{jumping-falling-pickup}
        \label{fig:figureS5a}
    \end{subfigure}
    \hfill
    \begin{subfigure}[h]{0.32\textwidth}
        \centering
        \includegraphics[width=\textwidth]{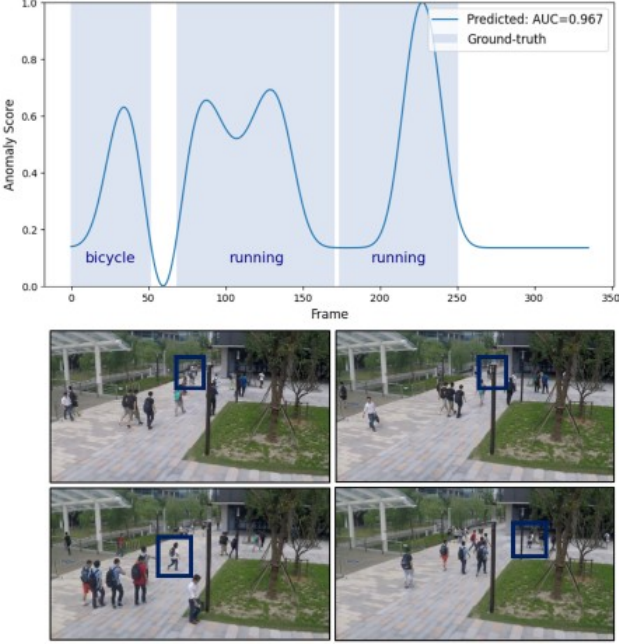}
        \caption{bicycle-running}
        \label{fig:figureS5b}
    \end{subfigure}
    \hfill
    \begin{subfigure}[h]{0.32\textwidth}
        \centering
        \includegraphics[width=\textwidth]{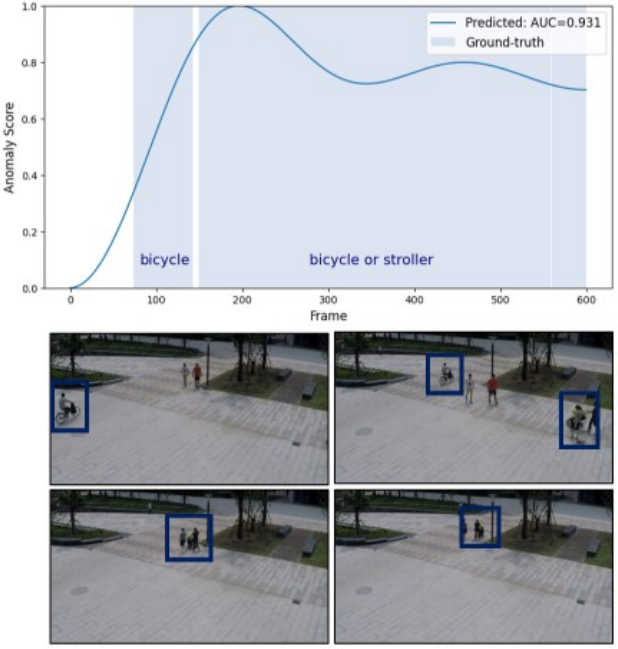}
        \caption{bicycle-stroller}
        \label{fig:figureS5c}
    \end{subfigure}
    \caption{Anomaly detection in diverse scenarios. Various abnormal events can emerge over time.}
    \label{fig:figureS5}
\end{figure*}

\subsection{Context Complementarity}
\label{supply3.2}
In this section, we explain the complementarity between $PC$ and $TC$ in context-aware VQA. \Cref{fig:figureS4} visualizes the key frame of a specific segment along with the images generated using WA and GIG for of $PC$ and $TC$. We also present the results of a context-aware VQA that utilizes these contexts.

In the first row, when the text input was `bicycle', $PC$ successfully identified the bicycle via WA, yielding a score of 1.0. However, the temporal context suffers from a cropping effect due to motion over time, resulting in a lower score of 0.5. In the second row, when the text input is `jumping,' the attention result from WA fails to accurately locate the `jumping' person. Additionally, because of the lack of temporal information, $PC$ was unable to recognize the jumping action, resulting in a score of 0.0. In contrast, $TC$ captured the entire jumping action over time, achieving a score of 0.9.

These results demonstrate that the proposed $PC$, which focuses on the object appearance, and $TC$, which leverages temporal information, are complementary. By integrating both approaches, we enable an effective generalization of the VAD.

\subsection{Anomaly Detection in Diverse scenarios}
\label{supply3.3}
\Cref{fig:figureS5} visualizes the results of VAD performed on videos containing multiple abnormal classes. The captions in each figure indicate the abnormal classes used in the corresponding video. We input the user-defined abnormal keywords as text individually to obtain the scores, and assigned the highest score as the anomaly score for the corresponding segment. As shown in the visualization results, the proposed AnyAnomaly enables VAD across various types of abnormal events. This demonstrates that AnyAnomaly can be effectively utilized even when the user aims to simultaneously detect multiple abnormal types.

\subsection{Anomaly Detection in Complex scenarios}
\label{supply3.4}
\begin{figure*}[t]
    \centering
    \begin{subfigure}[h]{0.85\linewidth}
        \centering
        \caption{Anomaly event: jaywalking}
        \includegraphics[width=\textwidth]{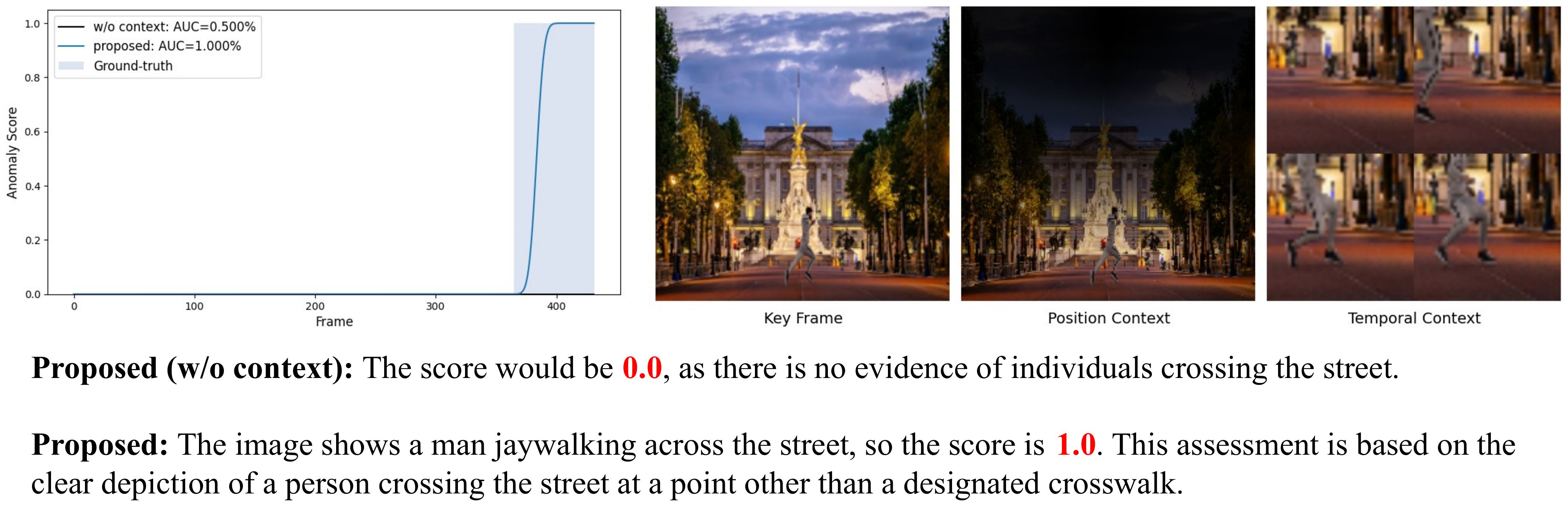}
        \label{fig:figureS6a}
        \vspace{-10pt}
    \end{subfigure}
    \begin{subfigure}[h]{0.85\linewidth}
        \caption{Anomaly event: driving outside lane}
        \includegraphics[width=\textwidth]{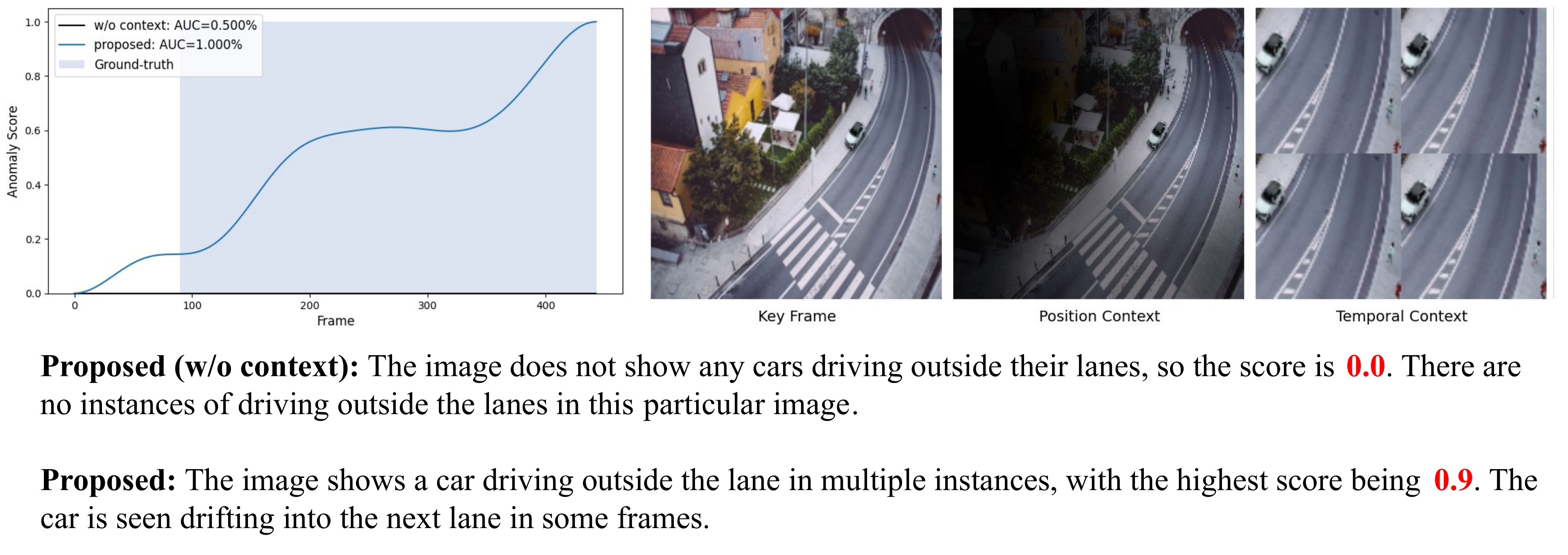}
        \label{fig:figureS6b}
        \vspace{-10pt}
    \end{subfigure}
    \begin{subfigure}[h]{0.85\linewidth}
        \caption{Anomaly event: people and car accident}
        \includegraphics[width=\textwidth]{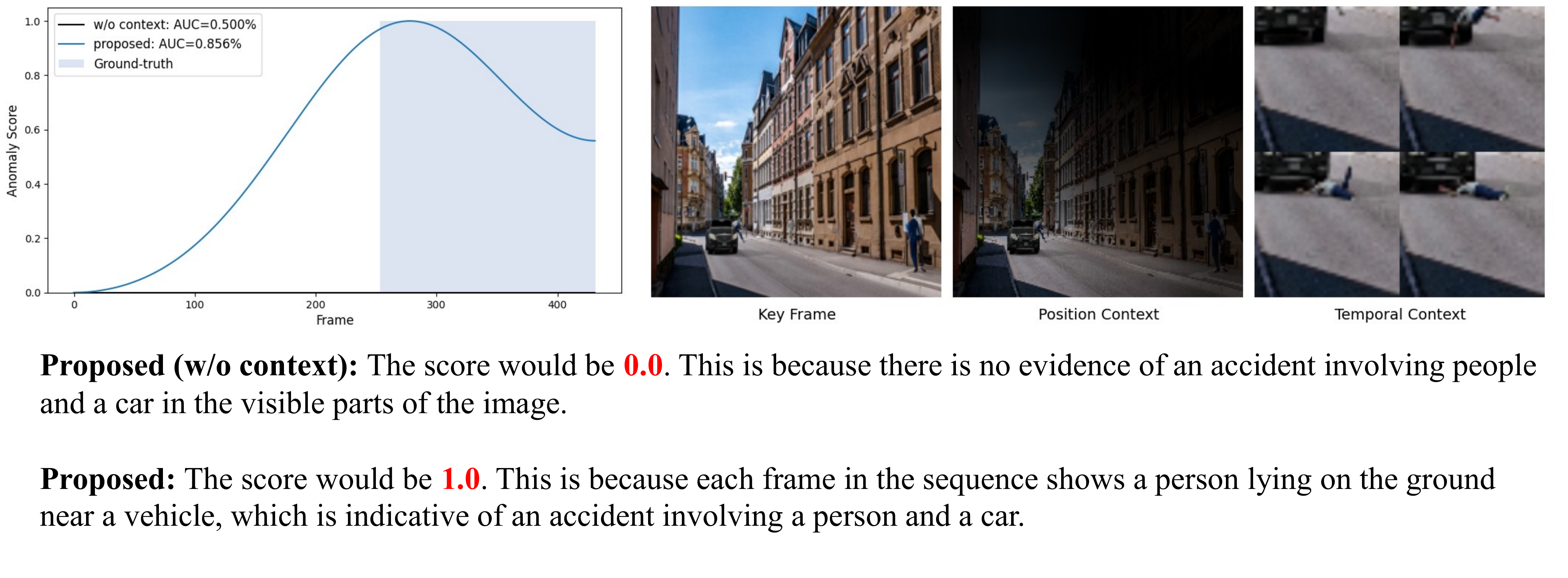}
        \label{fig:figureS6c}
        \vspace{-15pt}
    \end{subfigure}
    \begin{subfigure}[h]{0.85\linewidth}
        \caption{Anomaly event: walking drunk}
        \includegraphics[width=\textwidth]{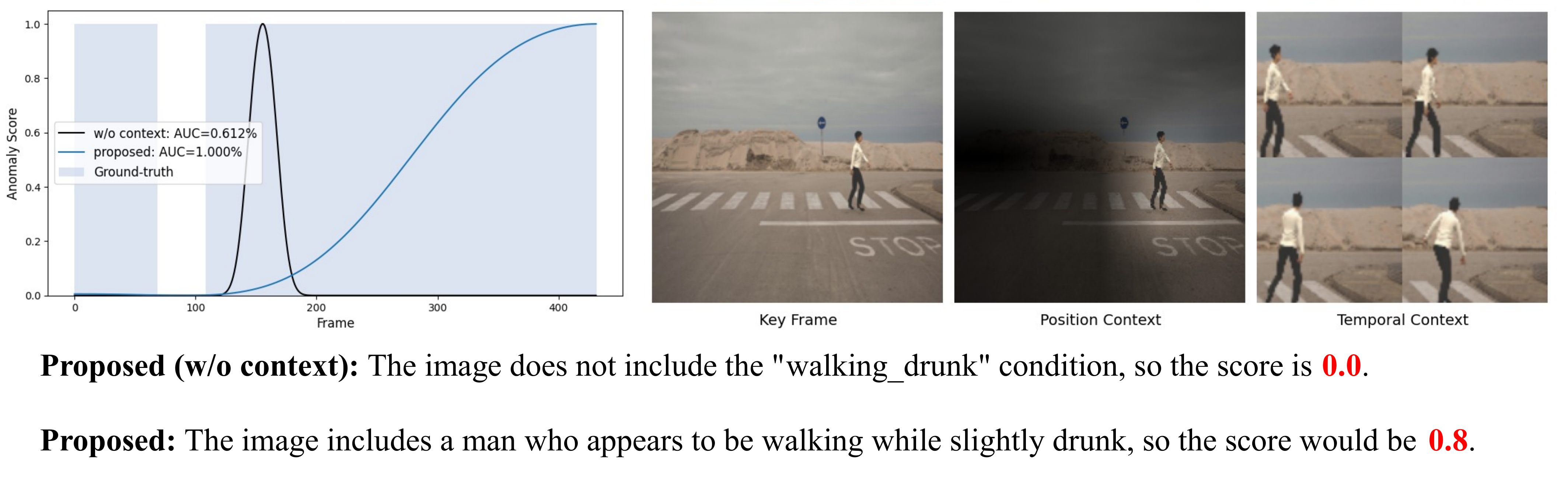}
        \label{fig:figureS6d}
        \vspace{-20pt}
    \end{subfigure}
    \caption{Anomaly detection in complex scenarios. Results with and without the inclusion of context are presented.}
    \label{fig:figureS6}
\end{figure*}

\Cref{fig:figureS6} presents the visualization results of AnyAnomaly on complex scenarios. `Key Frame', `Position Context', and `Temporal Context' visualize $\hat{k}$, $PC$, and $TC$, respectively. The text below each figure represents the LVLM output. These visualization results demonstrate that the proposed context-aware VQA, which utilizes $PC$ and $TC$, is effective and contributes to improving VAD performance.

Additionally, in \Cref{fig:figureS6d}, we observe that the model can detect certain frames of ``walking drunk" even without utilizing context information. This suggests that the strong visual reasoning capabilities of the LVLM enable VAD in complex scenarios. However, as shown in \Cref{fig:figureS6a}--\ref{fig:figureS6c}, relying solely on individual frames is insufficient for fully leveraging these reasoning abilities. Therefore, the proposed context-aware VQA approach is essential for effective VAD.

\section{Discussion}
\subsection{Comparison with traditional VAD}
\label{supply4.1}
Traditional VAD methods and our zero-shot C-VAD each have distinct strengths and limitations. Traditional VAD detects anomalies as deviations from learned normal patterns, requiring no prior knowledge of specific anomaly types and delivering strong performance within the trained environment. However, it often exhibits poor generalization to unseen environments and typically necessitates retraining. In contrast, C-VAD requires prior knowledge of anomaly types but removes the need for retraining or additional data collection even when the definition of “normal” varies across users or environments. This makes it a practical and cost-effective solution for real-world applications. We anticipate that, with continued advances in LVLM technology, the proposed C-VAD will become even more effective in the future.\\

\subsection{Limitation}
\label{supply4.2}
Efficiency is crucial in VAD; therefore, we adopted the most lightweight model among the SOTA LVLMs and employed a segment-level approach to reduce latency. However, when multiple abnormal events occur simultaneously, each event must be processed independently, which increases latency. In future work, we plan to improve the efficiency of C-VAD in handling multiple abnormal events concurrently.

\end{document}